\documentclass[11pt,a4paper]{article}
\usepackage[hyperref]{acl2021}
\usepackage{times}
\usepackage{latexsym}

\usepackage{microtype}

\aclfinalcopy %
\def\aclpaperid{2278} %

\usepackage{amsmath,amsfonts,bm}

\def\eqref#1{equation~\ref{#1}}

\def\1{\bm{1}}

\def\va{{\bm{a}}}

\def\vf{{\bm{f}}}

\def\vt{{\bm{t}}}

\def\vx{{\bm{x}}}
\def\vy{{\bm{y}}}

\def\mE{{\bm{E}}}

\def\mW{{\bm{W}}}
\def\mX{{\bm{X}}}
\def\mY{{\bm{Y}}}

\DeclareMathAlphabet{\mathsfit}{\encodingdefault}{\sfdefault}{m}{sl}
\SetMathAlphabet{\mathsfit}{bold}{\encodingdefault}{\sfdefault}{bx}{n}

\newcommand{\R}{\mathbb{R}}

\newcommand{\softmax}{\mathrm{softmax}}
\newcommand{\gelu}{\mathrm{gelu}}
\newcommand{\sigmoid}{\sigma}

\usepackage{booktabs}
\usepackage{xspace}
\usepackage{tabularx}
\usepackage{xhfill}
\usepackage{xcolor}
\usepackage{multirow}
\usepackage{mdframed}
\usepackage{tikz}
\usepackage{graphicx}
\usepackage{bbm}

\usepackage{mathtools}
\usepackage[colorinlistoftodos,prependcaption,textsize=tiny]{todonotes}
\usepackage{xargs}
\mathtoolsset{showonlyrefs}

\newcommandx{\rahul}[2][1=]{\todo[linecolor=red,backgroundcolor=red!25,bordercolor=red,#1]{#2}}
\newcommandx{\shashi}[2][1=]{\todo[linecolor=blue,backgroundcolor=blue!25,bordercolor=blue,#1]{#2}}

\definecolor{forestgreen}{HTML}{009B55}
\definecolor{sepia}{HTML}{671800}
\definecolor{midnightblue}{HTML}{006795}
\definecolor{orangered}{HTML}{E24C00}

\newcommand\gold{\textsc{Gold}\xspace}
\newcommand\roberta{\textsc{RobertaS2S}\xspace}
\newcommand\pegasus{\textsc{Pegasus}\xspace}
\newcommand\fame{\textsc{Fame}\xspace}
\newcommand\famep{\textsc{PegFame}\xspace}
\newcommand\famer{\textsc{RobFame}\xspace}
\newcommand\oracle{\textsc{Oracle}\xspace}

\newcommand\rouge{\textsc{rouge}\xspace}
\newcommand\rougeone{\textsc{rouge-1}\xspace}
\newcommand\rougetwo{\textsc{rouge-2}\xspace}
\newcommand\rougel{\textsc{rouge-l}\xspace}

\newcommand\xsum{\textsc{XSum}\xspace}
\newcommand\cnndm{\textsc{Cnn/Dm}\xspace}

\newcommand\ftopk{$\mathrm{Focus}_{\mathrm{top},k}$\xspace}
\newcommand\fsamplek{$\mathrm{Focus}_{\mathrm{sample},k}$\xspace}
\newcommand\divtopk{$\mathrm{Div}_{\mathrm{top},k}$\xspace}
\newcommand\divnucleus{$\mathrm{Div}_{\mathrm{nucleus}}$\xspace}

\newcolumntype{Y}{>{\centering\arraybackslash}X}

\title{Focus Attention: Promoting Faithfulness and Diversity in Summarization}

\author{
Rahul Aralikatte\thanks{~Work done when authors were interning/working at Google.} \\
University of Copenhagen \\ \texttt{rahul@di.ku.dk} \And
Shashi Narayan \\ Google Research \\ \texttt{shashinarayan@google.com} \AND 
Joshua Maynez \\ Google Research \\ \texttt{joshuahm@google.com} \And 
Sascha Rothe \\ Google Research \\ \texttt{rothe@google.com} \And
Ryan McDonald$^{*}$ \\ ASAPP \\ \texttt{ryanmcd@asapp.com}
}

\date{}

\begin{document}
\maketitle

\begin{abstract}
Professional summaries are written with document-level information, such as the theme of the document, in mind. This is in contrast with most seq2seq decoders which simultaneously learn to focus on salient content, while deciding what to generate, at each decoding step. With the motivation to narrow this gap, we introduce Focus Attention Mechanism, a simple yet effective method to encourage decoders to proactively generate tokens that are similar or topical to the input document. Further, we propose a Focus Sampling method to enable generation of diverse summaries, an area currently understudied in summarization. When evaluated on the BBC extreme summarization task, two state-of-the-art models augmented with Focus Attention generate summaries that are closer to the target and more faithful to their input documents, 
outperforming their vanilla counterparts on \rouge and multiple faithfulness measures. 
We also empirically demonstrate that Focus Sampling is more effective in generating diverse and faithful summaries than top-$k$ or nucleus sampling-based decoding methods.
\end{abstract}

\section{Introduction}
\label{sec:intro}

Document summarization --- producing the shorter version of a document while preserving salient information \cite{mani2001automatic,Nenkova:McKeown:2011} --- is challenging even for humans. Today, systems can generate summaries with a high level of fluency and coherence. This is due to recent advances such as sequence-to-sequence architectures (seq2seq) with attention and copy mechanism \cite{lstm,Bahdanau2015NeuralMT,gu-etal-2016-incorporating}, fully attention-based Transformer architectures \cite{transformer}, and large pretrained language models
\cite{bert,gpt,xlnet_arxiv19,roberta,unilm_arxiv19,mass_icml19,bart,rothe2020leveraging,t5,zhang2019pegasus}.

\begin{figure}[t!]
  \center{\scriptsize %
  \setlength\tabcolsep{0.1cm}
    \begin{tabular}{p{0.1cm}p{7.0cm}}
    \toprule 
    \multirow{9}{*}{A} & \textbf{\gold}: Australia has expelled an Israeli diplomat saying Israel was behind the forging of Australian passports linked to the murder of a Hamas operative in Dubai. \\
    & \textbf{\pegasus}: Australia has expelled an Israeli diplomat after concluding that forged Australian passports used in the killing of a Hamas militant in Dubai were issued by Israel.\\
    & \textbf{Our \famep model}: The Australian government has expelled an Israeli diplomat over the use of forged Australian passports in the killing of a Hamas militant in Dubai.\\
    \midrule
    \multirow{15}{*}{B} & \textbf{\pegasus with Top-$k$ Sampling} \\
    & \textcolor{orangered}{Israel has summoned the Australian ambassador to complain} after the Australian government said forged passports used in the killing of a Hamas operative in Dubai belonged to Netanyahu's foreign ministry. \\
    & The Australian government has ordered Israel to withdraw an officer over the use of forged Australian passports used \textcolor{orangered}{by the 2013 murder of a Lebanese opposition figure in Dubai.} \\
    & \textbf{\pegasus with Nucleus Sampling} \\
    & \textcolor{orangered}{Israel hasracuse withdrawn an envoy after} the Australian government said it concluded that Israeli agents used forged passports used to kill a Dubai \textcolor{orangered}{Bendigo businessman.}\\
    & The Australian government has recalled an Israeli diplomat over accusation that fake Australian passports \textcolor{orangered}{used 436 kilometres (300 miles) from Canberra} in the death of a Hamas militant \textcolor{orangered}{were stolen by Israeli agents.}\\
    \midrule
    \multirow{7}{*}{C} & \textbf{Our \famep model with novel Focus Sampling} \\
    & Australia has expelled an Israeli diplomatic staff after accusing the country's security agency, the Israeli military's intelligence agency, of being responsible for the use of Australian visas used in the killing of a Palestinian. \\
    & The Australian government has expelled an Israeli diplomatic staff after it said the country was responsible for the use of Australian visas used in the killing of a Palestinian in the Middle East. \\
    \bottomrule
    \end{tabular}     
  }
  \caption{Block A shows the best predictions from \pegasus and our \famep (\pegasus with \fame) model, along with the \gold summary for an \xsum article. Block B presents diverse summaries generated from \pegasus using top-$k$ and nucleus sampling. Block C shows diverse summaries generated using our \famep model with Focus sampling. The text in \textcolor{orangered}{orange} is not supported by the input article. 
  }%
  \label{fig:intro-fame-predictions}
\end{figure}

\noindent However, in terms of summary quality, many challenges remain. For example, generating summaries that are faithful to the input is an unsolved problem \cite{kryscinski-etal-2020-evaluating,maynez2020faithfulness,gabriel2020figure}.
Furthermore, there can be multiple equally good summaries per source document. Neural generation models fail to account for this and
tend to generate outputs with low diversity due to standard likelihood training, approximate decoding objectives, and lack of high quality multi-reference datasets \cite{fan-etal-2018-hierarchical,kulikov-etal-2019-importance,freitag-etal-2020-bleu,choi-etal-2020-f}. Not much attention has been given to generation of diverse, yet faithful summaries -- two goals are often challenging to achieve simultaneously \cite{hashimoto-etal-2019-unifying}; a model can produce  diverse outputs through sampling \cite{fan-etal-2018-hierarchical,nucleus}, but at the cost of quality.

In this paper we introduce a Focus Attention MEchanism (or \fame) to transformer-based seq2seq architectures. \fame is inspired by how humans write summaries. Specifically, \fame aims to perform source-side planning to focus the summary on supported and topical content.
\fame achieves this through a novel technique which augments standard contextual representations with a dynamic source-conditioned vocabulary biasing layer.
We present the following experimental findings:

\paragraph{\fame promotes summaries faithful to the source}
When evaluated on the BBC extreme summarization task (\xsum; \citeauthor{narayan-etal-2018-xsum}, \citeyear{narayan-etal-2018-xsum}), experiments with two state-of-the-art summarizers -- \roberta \cite{rothe2020leveraging} and \pegasus \cite{zhang2019pegasus} --
show that both models generate summaries that are more faithful to their input documents when augmented with \fame, in comparison with their vanilla counterparts.\footnote{In the paper we focus on assessing \fame on \xsum. But other summarization and text editing results can be found in Appendix \ref{sec:cnn-results} and \ref{sec:texted-results}.}
Faithfulness is measured through a variety of previously proposed metrics. In addition, we leverage the manually annotated document-summary pairs for faithfulness from \newcite{maynez2020faithfulness} and train a scorer which serves as an efficient proxy for expensive human evaluations. We call this metric {\em BERTFaithful}.

\paragraph{\fame enables diverse summaries}
\fame, by design, supports {\em Focus Sampling} -- a technique that is more effective in sampling topically relevant tokens to generate diverse, yet topically consistent and faithful outputs, than other sampling methods \cite{fan-etal-2018-hierarchical,nucleus}. Figure~\ref{fig:intro-fame-predictions} illustrates how focus sampling generates better summaries than other sampling methods. We demonstrate the effectiveness of our new Focus Sampling technique using a variety of existing diversity and faithfulness measures. Empirically, we find that optimizing for high diversity often comes at the cost of faithfulness. Thus \fame provides a mechanism for trading-off high faithfulness with better diversity in summarization.

\section{Related Work}
\label{sec:related}

\paragraph{Task-Specific Architectural Priors}
Several works enhance seq2seq architectures with task-specific priors. Pointer-generator style models \cite{see-acl17,xu-etal-2020-self} can accurately generate mostly extractive summaries by copying words from the source text via pointing. Text editing models \cite{malmi-etal-2019-encode,dong-etal-2019-editnts,mallinson2020felix} cast text generation as a sequence tagging problem with carefully selected edit operations required for the task. Others focus on improving content selection to better constrain the model to likely input phrases \cite{gehrmann-emnlp18} or by improving the representation of relevant input tokens \cite{zhou-etal-2017-selective}. Instead of directly modeling such priors, \fame learns the  theme of the document through dynamic vocabulary biasing. Thus, \fame can be seen as a generalization of Pointer-generator or text-editing models via soft vocabulary learning. In fact, our \fame models achieve state-of-the-art on text-editing tasks (Appendix~\ref{sec:texted-results}).

\paragraph{Topic-Aware Generation Models}
The idea of capturing document-level semantic information has been widely explored in the summarization community. \newcite{barzilay-elhadad-1997-using} use WordNet \cite{fellbaum98wordnet} to model a text's content relative to a topic based on lexical chains. \newcite{lin-hovy-2000-automated} propose to learn topic signatures for summarizing documents. Recently, document-level topic information has been used for improving neural language models \cite{mikolovZ12,Ghosh2016ContextualL,dieng-iclr17,karmaker-santu-etal-2019-tilm}, neural response generators \cite{topicgen-aaai17,dziri-etal-2019-augmenting}, and not surprisingly, neural summarizers \cite{narayan-etal-2018-xsum,DBLP:journals/corr/abs-1908-07026,wang-etal-2020-friendly}. Both, \newcite{narayan-etal-2018-xsum} and \newcite{DBLP:journals/corr/abs-1908-07026}, use a pretrained Latent Dirichlet Allocation (LDA; \citeauthor{Blei:2003:LDA}, \citeyear{Blei:2003:LDA}) model, whereas, \newcite{wang-etal-2020-friendly} use Poisson factor analysis \cite{pmlr-v22-zhou12c}, to synthesize topic vectors for the input. Instead, we dynamically learn a target-induced topic distribution for the input under the assumption that the human-written summary is a good proxy for the input document.

\paragraph{Faithful Generation Models}
\newcite{cao2017faithful} force faithful generation by conditioning on both source text and extracted fact descriptions from the source text. \newcite{Song2020JointPA} propose to jointly generate a sentence and its syntactic dependency parse to induce grammaticality and faithfulness. \newcite{tian2019sticking} learn a confidence score to ensure that the model attends to the source whenever necessary. \newcite{wang2020faithful} introduce new input-output matching and embedding similarity losses to alleviate hallucination issues. 
Yet, the task of generating text that is consistent with the input remains an open problem \cite{gabriel2020figure}.

\paragraph{Diverse Generation Models}
There has been a surge of interest in making language models generate more diverse and human-like outputs. \newcite{diversebeam} and \newcite{kulikov-etal-2019-importance} diversify beam search, using a task-specific scoring function, or constrain beam hypotheses to be sufficiently different. Others avoid text degeneration by truncating the unreliable tail of the probability distribution at each decoding step, either by sampling from the top-$k$ tokens ({\em Top-$k$ Sampling}; \citeauthor{fan-etal-2018-hierarchical}, \citeyear{fan-etal-2018-hierarchical}) or by sampling from a dynamic nucleus of tokens with the bulk of the probability mass ({\em Nucleus Sampling}; \citeauthor{nucleus}, \citeyear{nucleus}). Others modify the training objective to make the distribution sparse \cite{martins-etal-2020-sparse} or assign lower probability to unlikely generations \cite{Welleck2019}. 
 
For conditional text generation, most work focuses on generating diverse questions \cite{narayan-etal-2016-paraphrase,dong-etal-2017-learning,sultan-etal-2020-importance,wang-etal-2020-diversify} or paraphrases \cite{LiMJ16,DaiLUF17,xu-etal-2018-diversity,cao-wan-2020-divgan}. Following \newcite{gehrmann-emnlp18}, \newcite{cho-etal-2019-mixture} use a mixture of experts to sample different binary masks on the source sequence for diverse content selection for summarization.

\noindent Our focus sampling is similar to top-$k$ and nucleus sampling methods; in that it truncates the tail of the probability distribution. However, instead of truncating it at each decoding step, it biases the decoder proactively to generate output from a set of tokens which are topically-relevant to the input.

\begin{figure}[t!]
    \centering
    \includegraphics[width=\columnwidth]{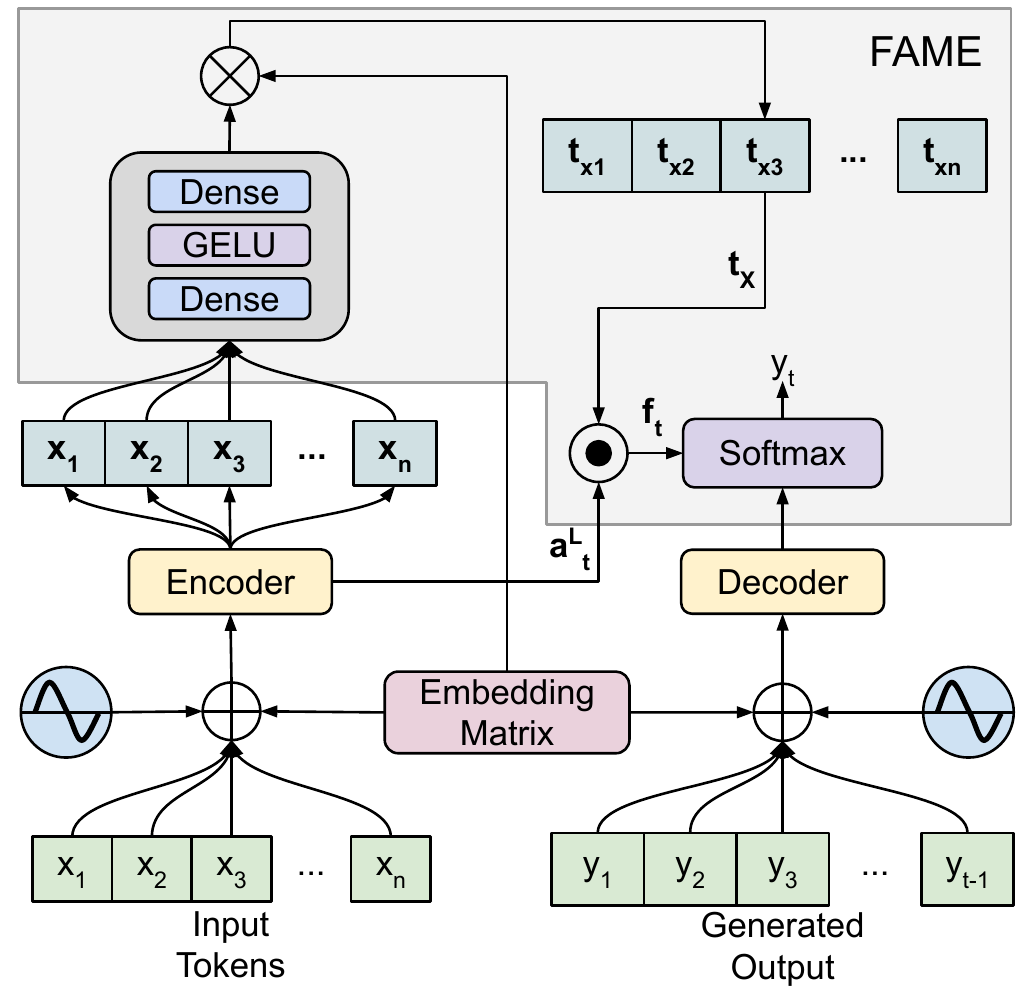} %
    \caption{A Transformer-based encoder-decoder architecture with \fame. }
    \label{fig:model}
\end{figure}

\section{Summarization with Focus Attention}
\label{sec:model}

Given an input document $X_{1:n}$, we aim to generate its summary $Y_{1:m}$, where $n$ and $m$ are input and output sequence lengths. We address this problem using seq2seq architectures with Transformer encoder and decoder, augmented with \fame, as depicted in Figure~\ref{fig:model}.
\fame learns a distribution $\vt_{x_i}$ for each input token $x_i$ over the vocabulary, measuring similarity of $x_i$ (in context) to the tokens in the vocabulary. The vocabulary distributions, $\vt_{x_i}$, for all  $x_i$ are combined to form a dynamic vocabulary bias that is added to the decoder logits. This mechanism enhances the conditioning on the input source and encourages the decoder to generate tokens that are topically similar to the input.

\paragraph{Transformer-based seq2seq Model}
The encoder uses BERT Transformer layers with multi-headed self-attention to encode $X$ to a vector sequence $\mX = \vx_1, \ldots, \vx_n$, with $\vx_i \in \R^h$, where $h$ is the size of hidden representation. The decoder uses an identical architecture, except that
at decoding step $t$, layer $l$ adds a conditional representation $\vy^l_t \in \R^h$ for the token $y_t$ by attending to the output representation $\mY^{l-1}_{1:t-1} = \vy^{l-1}_1, \ldots, \vy^{l-1}_{t-1}$ generated so far through self-attention and by attending to the input contextual representation $\mX$ through encoder-decoder attention. The probability of predicting the next token $y_t$ from a vocabulary $V$ is: 
\begin{equation}
p(y_t|Y_{1:t-1},X; \theta) = \softmax(\mE \vy^L_t), \label{eq:basic}
\end{equation}
where, $\vy^L_t$ is the representation from the final decoder layer $L$, $\mE \in \R^{|V| \times h}$ the embedding matrix and $\theta$ the model parameters. Parameters are trained by minimizing cross-entropy at each decoding step: 
\begin{equation}
L_{\mathrm{MLE}}(\theta) = - \frac{1}{m} \sum^m_{i=1} \log p(\hat{y}_t|\hat{Y}_{1:t-1},X; \theta), \label{eq:mle}
\end{equation}
where, $\hat{Y}_{1:m}$ is the human-written summary.

\paragraph{Focus Attention MEchansim (\fame)} 

It is challenging for a decoder to obtain all relevant information from the conditional representation $\vy^L_t$ to learn the vocabulary output logits such that predictions $y_t$ are consistent with the input. 
Other modeling factors, specifically the decoder language model, can overwhelm model predictions.
\fame (Figure~\ref{fig:model}) addresses this by introducing a short-circuit from the source to the vocabulary output logits via a source-conditioned bias on vocabulary items.

We take the encoder representation $\mX = \vx_1, \ldots, \vx_n$ and learn a \emph{Token-level Vocabulary Distribution} $\vt_{x_i} = \gelu(\vx_i \mW_1) \mW_2 \mE \in \R^{|V|}$, for each token $x_i$ in the input sequence $\mX$. $\vt_{x_i}$ measures the contextual similarity of the input token $x_i$ to the tokens in the vocabulary; $\mW_1 \in \R^{h \times h'}$ and $\mW_2 \in \R^{h' \times h}$ are parameters of newly introduced dense layers, $h'$ is the intermediate filter size. We define a \emph{Source-conditioned Vocabulary Distribution} as $\vt_X = 1/n \sum^n_{i=1} \vt_{x_i} \in \R^{|V|}$ as an average of token-level vocabulary distributions for tokens present in the input sequence $X$, capturing the similarity of $X$ to the tokens in the vocabulary. 

Let $\va^L_t \in \R^n$ be the encoder-decoder attention distribution over the source tokens for the output token $y_t$ and the final decoder layer $L$. We use $\va^L_t$ to produce a weighted sum of the token-level vocabulary distributions to compute a dynamic vocabulary bias, or \emph{Focus Bias} $\vf_t = \sum^n_{i=1} \va^L_{t,i} \vt_{x_i} \in \R^{|V|}$ at decoding step $t$. We modify the probability of predicting the next token $y_t$ from a vocabulary $V$ as: 
\begin{equation}
p(y_t|Y_{1:t-1},X; \theta) = \softmax(\vy^L_t \mE + \vf_t) \label{eq:focus}
\end{equation}
We call this \emph{Focused Probability Distribution}, and it modifies the output logits dynamically to put more focus on those tokens in the vocabulary which are similar to the attended tokens in $X$. The focus bias introduces a human-inspired control to the model where we do not generate the output in a fully abstractive manner (as in Eq.~\eqref{eq:basic}), but we proactively generate output tokens that are similar to the input tokens (as in Eq.~\eqref{eq:focus}).

\paragraph{Summary-induced Topic Focused Distribution}

We aim to guide our focus bias $\vf_t$ to be a better representative of the topical content relevant for the task. We achieve this by using the human-written summary $\hat{Y}$ as a proxy for the topical content of the input and impose the following prior on the source-conditioned vocabulary distribution $\vt_X$:
\begin{align}
L_{\mathrm{Topic}}(\theta) = - & \frac{1}{|V|} \sum^{|V|}_{i=1} ( [v_i \in \hat{Y}] \log (\sigmoid(\vt_{X,i})) \\ & + [v_i \notin \hat{Y}] \log (1-\sigmoid(\vt_{X,i})) ). \label{eq:losstopic}
\end{align}
We further refine Eq.~\eqref{eq:losstopic}  by replacing $\hat{Y}$ with $\hat{Y}_c = \hat{Y} - F$, where $F$ is a set of $|F|$ most frequent tokens in the vocabulary,\footnote{which are usually articles or other function words.} to improve focus on 
content words. Our final loss function is then
\begin{equation}
L = \lambda L_{\mathrm{MLE}} + (1-\lambda) L_{\mathrm{Topic}}, \label{eq:lossfinal}
\end{equation}
where, $\lambda$ is an hyper parameter.\footnote{$\lambda$ is set to 0.5 for all experiments.}

By enforcing $\vt_X$ to be a topic distribution for the input $X$, we encourage the focus bias $\vf_t$ to promote topically relevant tokens, and subsequently generate topically consistent outputs. Importantly, our focus bias with target-induced topic distribution is task-agnostic and less vulnerable to reference divergence issues \cite{dhingra-etal-2019-handling,maynez2020faithfulness}, and can learn any property embodied in the target relevant for the task. For example, depending on the task, $\vf_t$ can learn to favour input tokens (e.g., for mostly extractive summaries) or new tokens (e.g., for mostly abstractive summaries). This is in sharp contrast to models that introduce task-specific priors, e.g., the pointer-generator network \cite{see-acl17} that can copy words from the source text, but does not do well on extreme summarization which is highly abstractive in nature \cite{narayan-etal-2018-xsum}.

\paragraph{Focus Sampling: Promoting Diversity in Faithful Generation}
We introduce \emph{Focus Sampling} with \fame to construct a subset $V_k \subseteq V$ by 
sampling $k$ tokens from the topic distribution $\vt_X$ (\fsamplek). 
Then, we modify Eq.~\eqref{eq:focus} as 
\begin{align}
p(y_t| & Y_{1:t-1},X; \theta) = \\ &
\begin{cases}
    \softmax(\vy^L_t \mE + f_t)_i & \text{if } v_i \in V_k \cup F\\
    0,              & \text{otherwise.}
\end{cases}
\label{eq:diverse}
\end{align}
For document summarization, the subset $V_k$ will capture topically salient tokens necessary to generate a summary; $F$ is always added to $V_k$ to ensure that the model has access to function words. By tuning the parameters of sampling, we can enforce the model to control the faithfulness or diversity of the outputs.

Focus sampling has similarities to top-$k$ (\divtopk; \citeauthor{fan-etal-2018-hierarchical}, \citeyear{fan-etal-2018-hierarchical}) and nucleus sampling (\divnucleus; \citeauthor{nucleus}, \citeyear{nucleus}); in that they all aim to promote diversity. At each decoding step, the top-$k$ sampling diversifies the generation process by sampling a token from the top $k$ tokens in the final output distribution. Similarly, nucleus sampling samples from a dynamic nucleus of tokens containing the vast majority (with a cumulative probability $p$) of the probability distribution. Both top-$k$ and nucleus sampling shorten the tail of the output distribution at each decoding step, whereas focus sampling constrains the decoder to use a fixed and topically relevant vocabulary $V_k$. Unlike the other two techniques, \fsamplek can also benefit from standard beam search decoding, leading to superior generation that is not only diverse, but also consistent with the input document.

\section{Experimental Setup}
\label{sec:experiments}

In this section we present our experimental setup to assess the ability of our \fame models to generate faithful summaries and to demonstrate that focus sampling is  more  effective  in  generating  diverse and faithful summaries than other sampling-based decoding methods.

\subsection{Extreme Summarization}
\label{subsec:gentasks}

We evaluate \fame models on extreme document summarization (\xsum; \citeauthor{narayan-etal-2018-xsum}, \citeyear{narayan-etal-2018-xsum}). The \xsum summaries, are extreme in that the documents are summarized into single-sentence summaries. These summaries demonstrate a high level of abstractiveness, and generating them automatically requires document-level inference, abstraction, and paraphrasing. Due to their extreme nature, \xsum summaries are ideal to evaluate \fame models' ability to capture the theme of the document.\footnote{We further experiment with long-form story highlight generation (\cnndm; \citeauthor{hermann-nips15}, \citeyear{hermann-nips15}) and two text editing tasks: Sentence Fusion \cite{discofuse} and Sentence Splitting \cite{wikisplit}. Their results can be found in Appendix \ref{sec:cnn-results} and \ref{sec:texted-results}. Our \fame models achieve SOTA on both text-editing tasks.} We use on the original cased version consisting of
204,045/11,332/11,334 training/validation/test document-summary pairs. During training, the input documents are truncated to 512 tokens. The length of the summaries
are limited to 64.

\subsection{Pretrained Models with \fame}
\label{subsec:famemodels}

We introduce \fame to two popular seq2seq architectures: RoBERTa initialized seq2seq (\roberta,  \citeauthor{rothe2020leveraging}, \citeyear{rothe2020leveraging}) and \pegasus \cite{zhang2019pegasus}. We refer \roberta models with \fame as \famer and \pegasus with \fame with \famep.

We experiment with \roberta-Large with shared encoder and decoder; it has 24 layers, a hidden size of 1024, filter size of 4096, 16 attention heads, and a vocabulary with 50K sentence pieces \cite{sentencepiece}. \roberta has around 455M parameters and \famer has an additional 8M parameters. 

The best-performing \pegasus model from \newcite{zhang2019pegasus} is not directly comparable with \roberta. It does not share the encoder and decoder, it only has 16 layers, a hidden size of 1024, filter size of 4096, 16 attention heads, with a total of 568M parameters, and it also uses a much larger vocabulary with 91K sentence pieces. Hence, we trained our own \pegasus model. We use the same architecture as \roberta and pretrain it on a mixture of C4 \cite{t5} and HugeNews \cite{zhang2019pegasus} datasets with the original objective of generating salient GAP-sentences. 

Our experiments focus on this newly trained \pegasus model which has same number of parameters and vocabulary as \roberta. But in contrast to \roberta, the encoder-decoder attention in \pegasus is pretrained. This allows us to analyse how focus attention affects  pretrained (\pegasus) vs randomly-initialized (\roberta) encoder-decoder attentions.\footnote{See Appendix \ref{sec:impl} for implementation details and hyperparameter settings.}

\begin{table*}[th!]
\centering
\footnotesize
\begin{tabular}{ r|ccc|c|cccc|ccc}
\toprule
\multirow{3}{*}{Models} & \multicolumn{3}{c|}{\multirow{2}{*}{Lexical Overlap (w/ ref)}} & \multirow{2}{*}{Sem. Sim.} & \multicolumn{4}{c|}{Faithfulness} & \multicolumn{3}{c}{others} \\
& & & & & \multirow{2}{*}{ent.} & \multirow{2}{*}{Feqa} & \multicolumn{2}{c|}{BERTFaithful} & \multirow{2}{*}{Len.} & \multirow{2}{*}{Rep.($\downarrow$)} & R1(P\%) \\
& R1 & R2 & RL & BERTSc. & & & \% & conf. & & & With doc. \\ 
\midrule
\roberta & 41.45 & 18.79 & 33.90 & 80.6 & 39.1 & 19.8 & 21.5 & 0.216 & 21.2 & 24.2 & 71.1 \\ 
\famer & 42.15 & 19.68 & 34.81 & 80.8 & 41.3 & 21.2 & 22.7 & 0.226 & 20.8 & 20.7 & 72.5 \\
\midrule
\pegasus & 44.85 & 22.26 & 37.03 & 81.7 & 43.6 & 24.5 & 27.0 & 0.263 & 21.1 & 6.0 & 73.8 \\
\famep & \textbf{45.31} & \textbf{22.75} & \textbf{37.46} & \textbf{81.9} & \textbf{44.8} & \textbf{24.8} & \textbf{27.3} & \textbf{0.269} & 20.8 & \textbf{5.3} & \textbf{74.3}  \\
\bottomrule
\end{tabular}
\caption{Abstractive Summarization results on \xsum test set comparing \fame models with their baselines. For all our models, we use standard beam decoding with a beam size of 4 to generate the single best summary for a document. Focus sampling is not used here. See Section~\ref{subsec:eval} for details on the evaluation metrics reported. Best number for each metric is \textbf{boldfaced}.}
\label{tab:xsum-results-full}
\end{table*}

\subsection{Evaluation Metrics}
\label{subsec:eval}

\paragraph{Lexical Overlap} We report
{\em \rouge} F1 scores \cite{rouge} against reference summaries; in particular, we report on \rougeone and \rougetwo for informativeness and \rougel for fluency.\footnote{We lowercased candidate and reference summaries and used \texttt{pyrouge} with parameters ``-a -c 95 -m -n 4 -w 1.2.''}

\paragraph{Semantic Similarity} We report {\em BERTScore} \cite{bertscore} which computes the contextual similarity between a candidate and its reference summary.

\paragraph{Faithfulness}
\rouge and BERTScore do not correlate well with faithfulness of the generated summaries \cite{maynez2020faithfulness}. Human evaluation is traditionally considered as the gold standard for measuring faithfulness. But recent research has shown that even human evaluation has shortcomings \cite{schoch-etal-2020-problem}. Moreover, it is prohibitively expensive. This has led to the proposal of meta-evaluation metrics for various generation tasks \cite{durmus2020feqa,factcc,bleurt,comet}.

We evaluate \fame models on semantic inference metrics such as textual entailment \cite{Pasunuru-multireward18,welleck-etal-2019-dialogue,falke-etal-2019-ranking,Kryscinski2019EvaluatingTF} and question answering \cite{arumae-liu-2019-guiding,Wang2020AskingAA}. In particular, we report the probability of a summary entailing ({\em ent.}) its input document \cite{maynez2020faithfulness} and QA-based {\em Feqa} scores \cite{durmus2020feqa}. For ent.\ scores, we train an entailment classifier by fine-tuning a BERT-Large pretrained model \cite{bert} on the Multi-NLI dataset \cite{mnli}. For Feqa, we use a fine-tuned BART \cite{bart} language model for question generation to generate questions from the summaries, and a BERT-base model fine-tuned on SQuAD \cite{squad} to answer the generated questions with input document as context.\footnote{We used the Feqa code available here: \url{https://github.com/esdurmus/feqa/}.}

In addition to {\em ent.} and {\em Feqa}, we train a scorer leveraging manually annotated document-summary pairs for faithfulness, as a surrogate for human evaluation and call this metric {\em BERTFaithful}.\footnote{A very similar scorer was used in the GEM benchmark \cite{gem2021} to identify and extract the subset with faithful reference summaries from the XSum dataset \cite{narayan-etal-2018-xsum}.}
In particular, we finetune a BERT-Base classifier on 500 manually annotated document and gold summary pairs for the XSum dataset from \newcite{maynez2020faithfulness} to predict whether a summary is faithful to the input document or not.\footnote{Out of 500, 90\% of the document-summary pairs were used for training and the rest 50 document-summary pairs were used for validation. We used the validation set to estimate Spearman’s correlation coefficients of different metrics with the human assessment for faithfulness. We found that both entailment scores ({\em ent.}) and {\em BERTFaithful} are moderately correlated with faithfulness with correlation coefficients of 0.4387 and 0.3889, respectively. As such, we believe that BERTFaithful works as an efficient proxy for expensive human evaluation for faithfulness for XSum summaries. More work is needed to understand if BERTFaithful generalizes to other datasets.} We report the percentage of summaries that were faithful ($\frac{1}{N} \sum_i \mathbbm{1}[p_i(\mbox{faithful}) > 0.5]$) and the model's confidence to generate faithful summaries ($\frac{1}{N} \sum_i p_i(\mbox{faithful})$); $N$ is the total number of examples in the test set.

\paragraph{Diversity}
We report the number of times (out of $n$), a model is able to generate a completely new summary ({\em Unique}), and {\em Distinct-N} \cite{li-etal-2016-diversity}, measuring the lexical diversity in the generated summaries. Distinct-N is estimated as the number of distinct $n$-grams of order $n$ divided by the total number of $n$-grams of the same order, in all generated summaries. 

Finally, we also report the average length of summaries ({\em Len.}), repetition errors ({\em Rep.}, estimated as the percentage of summaries with at least one repetition of rare or content words), and \rougeone precision against the input document ({\em R1, P\%}), to better understand their quality.

\section{Results}
\label{sec:results}

\begin{table*}[t!]
\centering
\footnotesize
\begin{tabular}{r|cccc|ccc|c|c}
\toprule
\multirow{2}{*}{Metrics} & \multirow{2}{*}{Unique} & \multicolumn{3}{c|}{Dist.-N} & \multicolumn{3}{c|}{\rouge} & \multirow{2}{*}{ent.} & \multirow{2}{*}{BERTSc.} \\
& & 1 & 2 & 3 & R1 & R2 & RL & \\
\midrule
\roberta (\divtopk) & 9.98 & 2.5 & 25.0 & 57.7 & 33.6 & 12.0 & 26.5 & 21.8 & 76.9 \\
\roberta (\divnucleus) & 9.99 & \textbf{4.1} & 30.1 & 62.2 & 32.4 & 11.4 & 25.6 & 19.7 & 75.7 \\
\midrule
\famer (\divtopk) & 9.99 & 2.3 & 25.0 & 58.1 & 32.7 & 11.3 & 25.7 & 20.3 & 76.6 \\
\famer (\divnucleus) & 9.99 & \textbf{4.1} & \textbf{30.7} & \textbf{63.2} & 31.3 & 10.6 & 24.7 & 18.0 & 75.4 \\
\famer (\fsamplek) & 1.61 & 3.5 & 22.4 & 43.9 & \textbf{38.0} & \textbf{15.7} & \textbf{31.0}  & \textbf{34.3} & \textbf{78.6} \\
\midrule
\famer (\fsamplek, \divtopk)& 9.99 & 2.1 & 20.3 & 51.8 & 31.8 & 10.2 & 24.7 & 24.3 &  75.4 \\
\famer (\fsamplek, \divnucleus)& 9.98 & 1.9 & 18.4 & 48.2 & 32.9 & 11.1 & 25.8 & 25.9 &  76.1 \\
\midrule\midrule
\pegasus (\divtopk) & 9.98 & 1.9 & 23.2 & 55.3 & 36.6 & 14.3 & 28.8 & 27.7 & 78.4  \\
\pegasus (\divnucleus) & 9.99 & \textbf{3.8} & \textbf{30.5} & \textbf{63.1} & 34.1 & 12.8 & 26.9 & 22.7 & 76.5  \\
\midrule
\famep (\divtopk)& 9.98 & 1.9 & 23.2 & 55.5 & 36.7 & 14.5 & 29.0 & 28.5 & \textbf{78.5}  \\
\famep (\divnucleus)& 9.99 & \textbf{3.8} & 30.4 & \textbf{63.1} & 34.2 & 12.8 & 27.0 & 23.2 & 76.6  \\
\famep (\fsamplek) & 2.77 & 2.4 & 16.5 & 34.2 & \textbf{37.5} & \textbf{15.4} & \textbf{30.3} & \textbf{33.6} & 77.9 \\
\midrule
\famep (\fsamplek, \divtopk) & 8.99 & 2.8 & 23.0 & 54.7 & 31.5 & 10.3 & 24.4 & 22.8 & 74.7 \\
\famep (\fsamplek, \divnucleus) & 9.98 & 2.6 & 20.8 & 50.9 & 32.5 & 11.0 & 25.3 & 24.8 & 75.3  \\
\bottomrule
\end{tabular}
\caption{Assessment of diversity, relevance and faithfulness with focus sampling on the \xsum test set.}
\label{tab:focus-sampling-quality}
\end{table*}

\paragraph{\fame Summaries are More Fluent, Informative and Faithful.} Table~\ref{tab:xsum-results-full} presents results comparing our \fame models, \famer and \famep, against their counterparts \roberta and \pegasus, respectively. Both \fame models clearly outperform their vanilla counterparts in terms of generating summaries that are more fluent (see RL and Rep.), more informative (see R1, R2 and BERTSc.) and more faithful (see ent., Feqa and BERTFaithful). Among all four models, \famep summaries are most fluent, informative and faithful. 

We further did pairwise comparisons for all measures in Table~\ref{tab:xsum-results-full} and found that all differences are statistically significant except for BERTScore and faithfulness measures between \pegasus and \famep.\footnote{All significance tests in this work are pairwise comparisons (one-way ANOVA with posthoc Tukey HSD tests; $p < 0.01$).} These assessments demonstrate that \fame models aid both \roberta and \pegasus in generating fluent, faithful and relevant summaries, but are more effective in \roberta than in \pegasus for extreme summarization.

\paragraph{Generating Diverse and Faithful Summaries with Focus Sampling.}
Table~\ref{tab:focus-sampling-quality} presents results assessing focus sampling (\fsamplek), top-$k$ sampling (\divtopk) and nucleus sampling (\divnucleus), for their abilities to generate diverse and faithful summaries. For \fsamplek, we choose $k=10,000$. %
We follow \newcite{nucleus} and choose $k=640$
and the nucleus probability $p=0.95$, for \divtopk and \divnucleus, respectively. For \fsamplek, we decode with a beam size of 4. We also report \fsamplek with \divtopk and \divnucleus to assess if they can benefit one-another.
In each setting we sample 10 summaries for each input document. 
For all metrics, we report the average over all 10 samples.\footnote{Feqa and BERTFaithful scores are dropped due to time constraints.}

Both \divtopk and \divnucleus almost always generate a new summary. In comparison \fsamplek generates 1.61 and 2.77 unique summaries using \famer and \famep models, respectively. \divnucleus tends to generate the most distinct unigrams, bigrams, and trigrams. Interestingly, \fsamplek summaries have a more diverse collection of unigrams than in \divtopk summaries (3.5\% vs 2.3\% for \famer and 2.4\% vs 1.9\% for \famep). 

The high diversity in \divtopk and \divnucleus comes at the cost of faithfulness; summaries generated with these sampling techniques have poor entailment scores. \fsamplek, on the other hand, generates summaries which entail documents the most. It also has the highest \rouge scores across the board. Some of the generated examples can be seen in Figure~\ref{fig:intro-fame-predictions}. More predictions from other models can be found in Appendix \ref{sec:diverse}. Augmenting \divtopk and \divnucleus with \fsamplek is not desirable because, though it increases diversity in terms of uniqueness and Distinct-$3$ scores, faithfulness suffers again.

Comparing results in Table~\ref{tab:focus-sampling-quality} to the results in Table~\ref{tab:xsum-results-full}, it is clear that diversity comes at the cost of quality (e.g., RL/ent. scores for \famer and \famer-\fsamplek are 34.81/41.3 and 31.0/34.3, respectively). However, \fsamplek is superior to both \divtopk and \divnucleus in generating better quality summaries.

\begin{figure}[th!]
    \centering
    \includegraphics[width=\columnwidth]{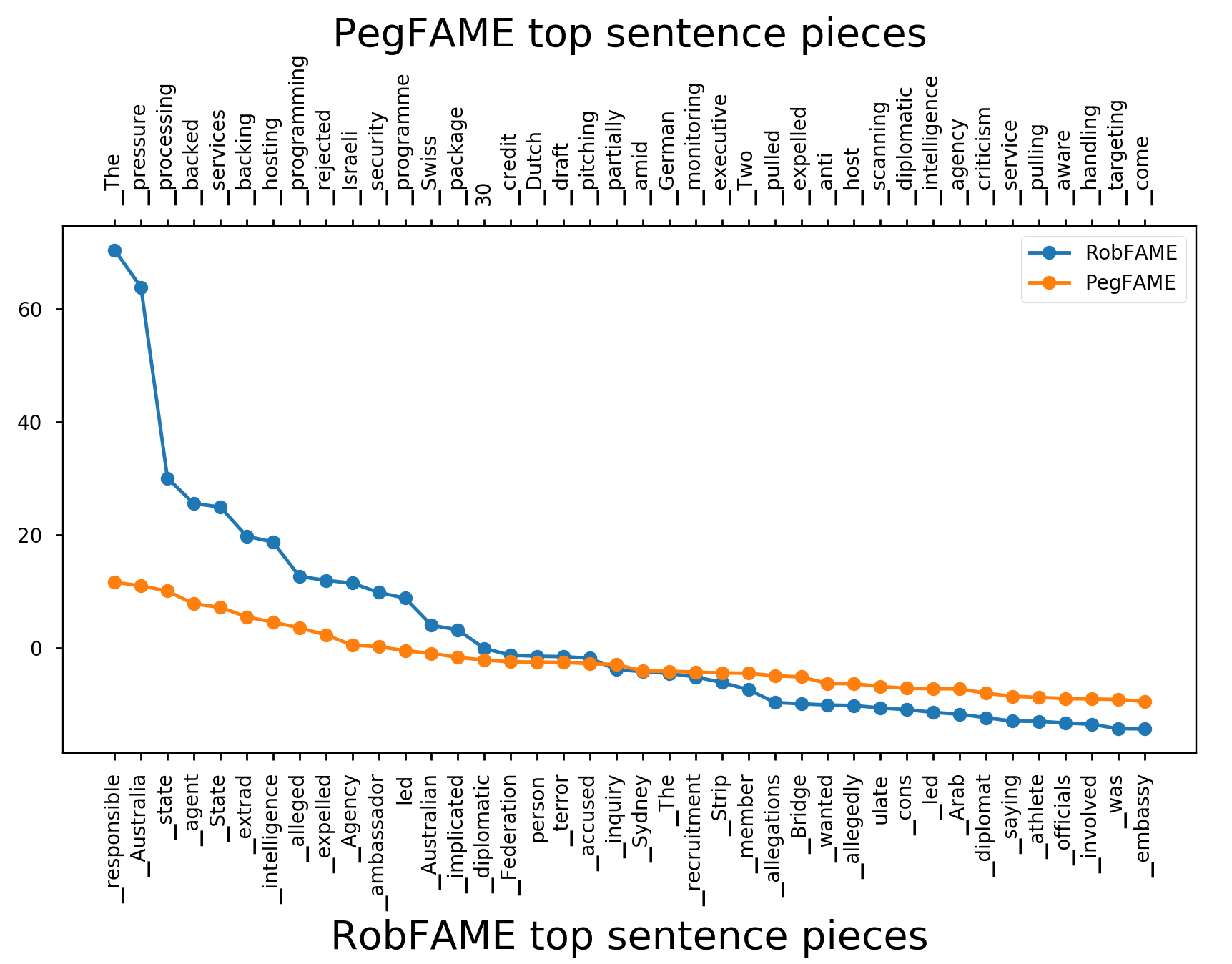}
    \caption{Top 40 sentence pieces and their logits from topic distribution $t_X$ in \famer and \famep for the \xsum article discussed in Figure~\ref{fig:intro-fame-predictions}. 
    }
    \label{fig:topk-sentpiece}
\end{figure}

\begin{figure}[th!]
    \centering
    \includegraphics[width=\columnwidth]{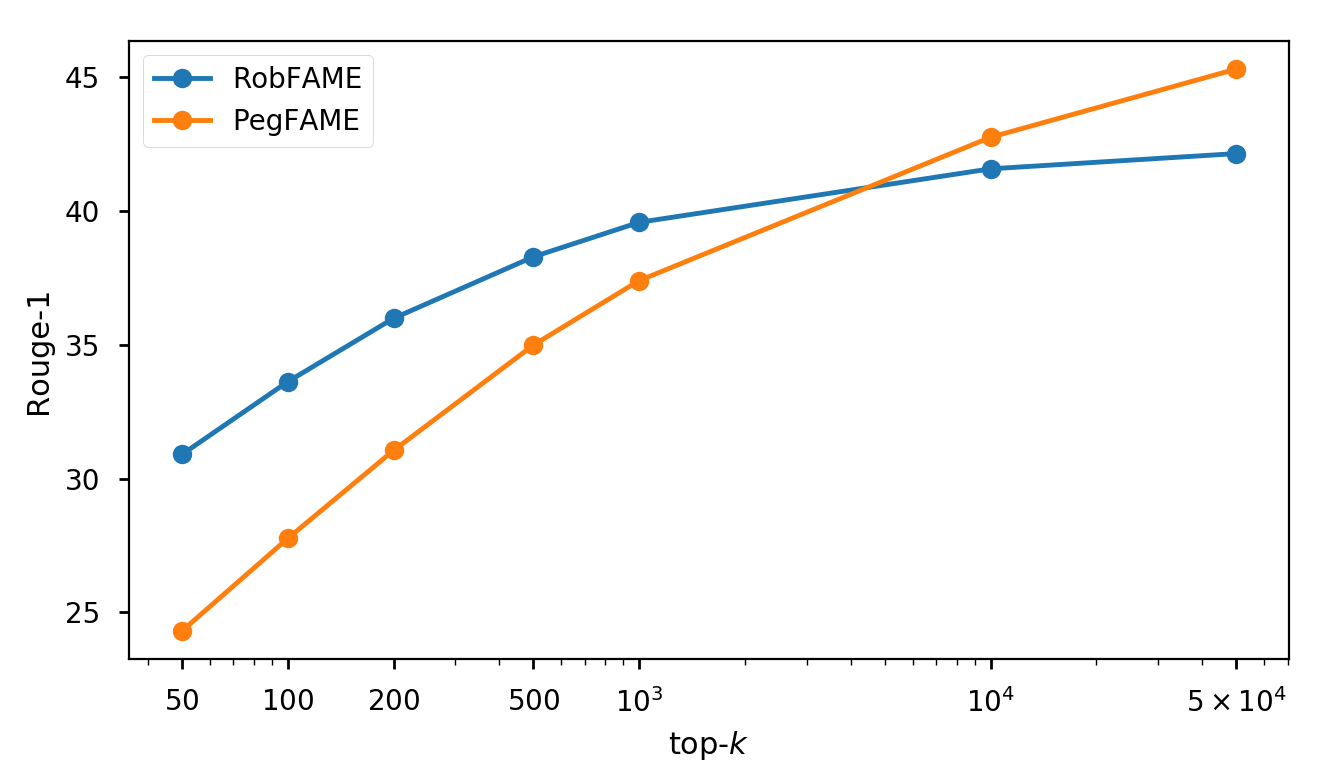}
    \caption{\rougeone F1 scores of \famer and \famep models with different top-$k$ vocabularies (Eq.~\eqref{eq:diverse}) on the \xsum test set. Similar patters are observed for \rougetwo and \rougel scores.}
    \label{fig:topk-rouge1}
\end{figure}

\paragraph{Focus Attention and Sampling Work Differently in \famer and \famep.}
Since both encoder-decoder and focus attention parameters of \famer are randomly initialized, they learn to compliment each other and learn a peaky topic distribution. On the other hand, since \famep's encoder-decoder attention is pre-trained, there is a push-pull effect between it and focus attention. This results in a smoother topic distribution, as seen in Figure \ref{fig:topk-sentpiece}.\footnote{This difference in topic distributions is consistent across the whole test set. We compute the peakiness score of a topic distribution as the slope of the line connecting logits of the top-1st token to the top-100th token. The average peakiness scores across the \xsum testset for \famer and \famep are 1.25 (51$^{\circ}$) and 0.45 (24.3$^{\circ}$), respectively.}

Although we see that both models' token sets capture the target intent well, the peaky distribution of \famer enables more accurate predictions than that of \famep, in a controlled generation setting. A comparison is presented in Figure~\ref{fig:topk-rouge1} where we show how \rougeone scores vary when we use only top-$k$ tokens from $t_X$ for generation.\footnote{Additional results and model predictions for these experiments can be found in Appendix \ref{sec:controlled}.} We observe that \famer consistently outperforms \famep with the lower values of $k\in \{50, 100, 200, 500, 1000\}$. 

\begin{table}[t!]
\centering
\scriptsize %
\begin{tabular}{r|ccc}
\toprule
Models & R1 & R2 & RL \\
\midrule
Lead & 16.30 & 1.61 & 11.95 \\
PtGen \cite{see-acl17} & 29.70 & 9.21 & 23.24  \\ 
ConvS2S \cite{narayan-etal-2018-xsum} & 31.89 & 11.54 & 25.75  \\
MMN \cite{Kim:2018:arXiv} & 32.00 & 12.10 & 26.00 \\ 
MASS \cite{mass_icml19} & 39.75 & 17.24 & 31.95\\
BART \cite{bart} & 45.14 & 22.27 & 37.25\\
\pegasus \cite{zhang2019pegasus} &  \textbf{\underline{47.21}} & \textbf{\underline{24.56}} & \textbf{\underline{39.25}}\\
\midrule
\roberta \cite{rothe2020leveraging} & 41.45 & 18.79 & 33.90  \\
\famer (w/o Eq.~\eqref{eq:losstopic}) & 41.27 & 18.86 & 33.90 \\
\famer & 42.15 & 19.68 & 34.81 \\
\oracle & \textbf{72.22} & \textbf{42.22} & \textbf{53.89} \\ \hline
\pegasus (ours) & 44.85 & 22.26 & 37.03 \\
\famep (w/o Eq.~\eqref{eq:losstopic}) & 44.54 & 22.00 & 36.83\\
\famep & 45.31 & 22.75 & 37.46 \\
\oracle & \textbf{82.39} & \textbf{60.61} & \textbf{69.19} \\
\bottomrule
\end{tabular}
\caption{Ablations and SOTA comparisons on \xsum dataset. The \textbf{\underline{underlined bold}} results are from the best performing models from literature and the \textbf{bold} results are the best performing \fame models.}
\label{tab:ablation-sota-results}
\end{table}

Further, we observe that \famer generates fewer unique summaries (1.61 vs 2.77) but has higher Distinct-N scores (3.5/22.4/43.9 vs 2.4/16.5/34.2) than \famep, with \fsamplek in Table~\ref{tab:focus-sampling-quality}. This can be again be attributed to how \fame works differently in \famer and \famep. When $V_k$ is sampled from \famer's peaky distribution, the beam search decoding often tends to generate similar summaries (leading to a lower Uniqueness score) as the sampled $V_k$s do not diverge by much from each other. But when it does diverge, the decoder tends to generate completely new summaries (leading to higher Distinct-N scores). 

Currently, we set $k=10,000$ for our focus sampling experiments following our observations in Figure~\ref{fig:topk-rouge1}. Future work will focus on how to better leverage trade-off between diversity and faithfulness by 
controlling the peakiness of the topic distribution $t_X$.

\paragraph{Ablations and SOTA Comparisons}

We emphasize that \fame or focus sampling does not aim to improve on state-of-the-results in terms of \rouge, but to generate more faithful or diverse summaries while maintaining their quality. For completeness, we compare our \famer and \famep models to their ablations and other state-of-the-art models on \xsum in Table~\ref{tab:ablation-sota-results}.

We report \rouge scores for \fame in the ideal scenario (\oracle) where it focuses on all the correct tokens in the input, i.e., the topic distribution $\vt_X$ is identical to the distribution observed in the reference summary.
These models generate summaries with very high \rouge scores when the model is given the correct tokens to focus on. The gap between the \oracle and \fame scores suggests that there is still a lot of work to be done in this space. 
Focus attention without any topical supervision (models w/o Eq.~\eqref{eq:losstopic}) is not significantly better than the baselines. But \famer and \famep (trained with joint supervision in Eq.~\eqref{eq:lossfinal}) significantly outperform \roberta and \pegasus, respectively.

Our best model \famep performs better than PtGen \cite{see-acl17},
ConvS2S \cite{narayan-etal-2018-xsum}, MMN \cite{Kim:2018:arXiv}, 
MASS \cite{mass_icml19} and BART \cite{bart}, but worse when the original \pegasus \cite{zhang2019pegasus}. This can be expected as the number of parameters in \famep is far less than that in the original \pegasus.

\section{Conclusion}
We introduced \fame, a new attention mechanism which dynamically biases the decoder to proactively generate tokens that are topically similar to the input. \fame enhances the faithfulness of existing state-of-the-art abstract summarization models while improving their overall \rouge scores. Finally, our newly introduced focus sampling technique is a better alternative to top-$k$ or nucleus sampling to generate diverse set of faithful summaries.

\section*{Acknowledgements}

We thank Sebastian Gehrmann, Slav Petrov, the reviewers, and the action editor for their invaluable feedback.

\section*{Ethical Considerations}

The nature of text generation leads to multiple ethical considerations when applied to applications. The main failure mode is that the model can learn to mimic target properties in the training data that are not desirable.
\paragraph{Faithfulness and Factuality} Since models create new text, there is the danger that they may neither be faithful to the source material nor factual. This can be exacerbated when the data itself has highly abstractive targets, which require the model to generate words not seen in the source material during training. This often leads the model to generate content inconsistent with the source material \cite{kryscinski-etal-2020-evaluating,maynez2020faithfulness,gabriel2020figure}.
\paragraph{Trustworthy Data} If the data itself is not trustworthy (comes from suspect or malicious sources) the model itself will naturally become untrustworthy as it will ultimately learn the language and topics of the training data. For instance, if the training data is about Obama birther conspiracies, and the model is asked to generate information about the early life of Obama, there is a risk that such false claims will be predicted by the model.
\paragraph{Bias in Data} Similarly, biases in the data around gender, race, etc., risk being propagated in the model predictions, which is common for most NLP tasks. This is especially true when the models are trained from non-contemporary data that do not represent current norms and practices \cite{blodgett-etal-2020-language}.

The above considerations are non-malicious, in that the model is merely learning to behave as its underlying source material. If users of such models are not aware of these issues and do not account for them, e.g., with better data selection, evaluation, etc., then the generated text can be damaging.
\\ \\
\noindent Generation models can also be misused in malicious ways. These include generating fake news, spam, and other text meant to mislead large parts of the general population.

\bibliography{ref}

\begin{thebibliography}{88}
\expandafter\ifx\csname natexlab\endcsname\relax\def\natexlab#1{#1}\fi

\bibitem[{Ailem et~al.(2019)Ailem, Zhang, and
  Sha}]{DBLP:journals/corr/abs-1908-07026}
Melissa Ailem, Bowen Zhang, and Fei Sha. 2019.
\newblock Topic augmented generator for abstractive summarization.
\newblock \emph{CoRR}, abs/1908.07026.

\bibitem[{Arumae and Liu(2019)}]{arumae-liu-2019-guiding}
Kristjan Arumae and Fei Liu. 2019.
\newblock Guiding extractive summarization with question-answering rewards.
\newblock In \emph{Proceedings of the 2019 Conference of the North {A}merican
  Chapter of the Association for Computational Linguistics: Human Language
  Technologies}, pages 2566--2577, Minneapolis, Minnesota.

\bibitem[{Bahdanau et~al.(2015)Bahdanau, Cho, and
  Bengio}]{Bahdanau2015NeuralMT}
Dzmitry Bahdanau, Kyunghyun Cho, and Yoshua Bengio. 2015.
\newblock Neural machine translation by jointly learning to align and
  translate.
\newblock \emph{CoRR}, abs/1409.0473.

\bibitem[{Barzilay and Elhadad(1997)}]{barzilay-elhadad-1997-using}
Regina Barzilay and Michael Elhadad. 1997.
\newblock \href {https://www.aclweb.org/anthology/W97-0703} {Using lexical
  chains for text summarization}.
\newblock In \emph{Intelligent Scalable Text Summarization}.

\bibitem[{Blei et~al.(2003)Blei, Ng, and Jordan}]{Blei:2003:LDA}
David~M. Blei, Andrew~Y. Ng, and Michael~I. Jordan. 2003.
\newblock Latent dirichlet allocation.
\newblock \emph{The Journal of Machine Learning Research}, 3:993--1022.

\bibitem[{Blodgett et~al.(2020)Blodgett, Barocas, Daum{\'e}~III, and
  Wallach}]{blodgett-etal-2020-language}
Su~Lin Blodgett, Solon Barocas, Hal Daum{\'e}~III, and Hanna Wallach. 2020.
\newblock \href {https://doi.org/10.18653/v1/2020.acl-main.485} {Language
  (technology) is power: A critical survey of {``}bias{''} in {NLP}}.
\newblock In \emph{Proceedings of the 58th Annual Meeting of the Association
  for Computational Linguistics}, pages 5454--5476, Online. Association for
  Computational Linguistics.

\bibitem[{Botha et~al.(2018)Botha, Faruqui, Alex, Baldridge, and
  Das}]{wikisplit}
Jan~A. Botha, Manaal Faruqui, John Alex, Jason Baldridge, and Dipanjan Das.
  2018.
\newblock Learning to split and rephrase from {W}ikipedia edit history.
\newblock In \emph{Proceedings of the 2018 Conference on Empirical Methods in
  Natural Language Processing}, pages 732--737, Brussels, Belgium. Association
  for Computational Linguistics.

\bibitem[{Cao and Wan(2020)}]{cao-wan-2020-divgan}
Yue Cao and Xiaojun Wan. 2020.
\newblock \href {https://www.aclweb.org/anthology/2020.findings-emnlp.218}
  {{D}iv{GAN}: Towards diverse paraphrase generation via diversified generative
  adversarial network}.
\newblock In \emph{Findings of the Association for Computational Linguistics:
  EMNLP 2020}, pages 2411--2421, Online. Association for Computational
  Linguistics.

\bibitem[{Cao et~al.(2017)Cao, Wei, Li, and Li}]{cao2017faithful}
Ziqiang Cao, Furu Wei, Wenjie Li, and Sujian Li. 2017.
\newblock \href {http://arxiv.org/abs/1711.04434} {Faithful to the original:
  Fact aware neural abstractive summarization}.

\bibitem[{Cho et~al.(2019)Cho, Seo, and Hajishirzi}]{cho-etal-2019-mixture}
Jaemin Cho, Minjoon Seo, and Hannaneh Hajishirzi. 2019.
\newblock \href {https://doi.org/10.18653/v1/D19-1308} {Mixture content
  selection for diverse sequence generation}.
\newblock In \emph{Proceedings of the 2019 Conference on Empirical Methods in
  Natural Language Processing and the 9th International Joint Conference on
  Natural Language Processing (EMNLP-IJCNLP)}, pages 3121--3131, Hong Kong,
  China. Association for Computational Linguistics.

\bibitem[{Choi et~al.(2020)Choi, Hong, Park, and Lee}]{choi-etal-2020-f}
Byung-Ju Choi, Jimin Hong, David Park, and Sang~Wan Lee. 2020.
\newblock \href {https://www.aclweb.org/anthology/2020.emnlp-main.737}
  {F{\^{}}2-softmax: Diversifying neural text generation via frequency
  factorized softmax}.
\newblock In \emph{Proceedings of the 2020 Conference on Empirical Methods in
  Natural Language Processing (EMNLP)}, pages 9167--9182, Online. Association
  for Computational Linguistics.

\bibitem[{Dai et~al.(2017)Dai, Lin, Urtasun, and Fidler}]{DaiLUF17}
Bo~Dai, Dahua Lin, Raquel Urtasun, and Sanja Fidler. 2017.
\newblock \href {http://arxiv.org/abs/1703.06029} {Towards diverse and natural
  image descriptions via a conditional {GAN}}.
\newblock \emph{CoRR}, abs/1703.06029.

\bibitem[{Devlin et~al.(2019)Devlin, Chang, Lee, and Toutanova}]{bert}
Jacob Devlin, Ming-Wei Chang, Kenton Lee, and Kristina Toutanova. 2019.
\newblock {BERT: P}re-training of deep bidirectional transformers for language
  understanding.
\newblock In \emph{Proceedings of the 2019 Conference of the North {A}merican
  Chapter of the Association for Computational Linguistics: Human Language
  Technologies}, pages 4171--4186, Minneapolis, Minnesota. Association for
  Computational Linguistics.

\bibitem[{Dhingra et~al.(2019)Dhingra, Faruqui, Parikh, Chang, Das, and
  Cohen}]{dhingra-etal-2019-handling}
Bhuwan Dhingra, Manaal Faruqui, Ankur Parikh, Ming-Wei Chang, Dipanjan Das, and
  William Cohen. 2019.
\newblock Handling divergent reference texts when evaluating table-to-text
  generation.
\newblock In \emph{Proceedings of the 57th Annual Meeting of the Association
  for Computational Linguistics}, pages 4884--4895, Florence, Italy.

\bibitem[{Dieng et~al.(2017)Dieng, Wang, Gao, and Paisley}]{dieng-iclr17}
Adji~B. Dieng, Chong Wang, Jianfeng Gao, and John Paisley. 2017.
\newblock {TopicRNN: A} recurrent neural network with long-range semantic
  dependency.
\newblock In \emph{Proceedings of the 5th International Conference on Learning
  Representations}, Toulon, France.

\bibitem[{Dong et~al.(2017)Dong, Mallinson, Reddy, and
  Lapata}]{dong-etal-2017-learning}
Li~Dong, Jonathan Mallinson, Siva Reddy, and Mirella Lapata. 2017.
\newblock \href {https://doi.org/10.18653/v1/D17-1091} {Learning to paraphrase
  for question answering}.
\newblock In \emph{Proceedings of the 2017 Conference on Empirical Methods in
  Natural Language Processing}, pages 875--886, Copenhagen, Denmark.
  Association for Computational Linguistics.

\bibitem[{Dong et~al.(2019{\natexlab{a}})Dong, Yang, Wang, Wei, Liu, Wang, Gao,
  Zhou, and Hon}]{unilm_arxiv19}
Li~Dong, Nan Yang, Wenhui Wang, Furu Wei, Xiaodong Liu, Yu~Wang, Jianfeng Gao,
  Ming Zhou, and Hsiao-Wuen Hon. 2019{\natexlab{a}}.
\newblock Unified language model pre-training for natural language
  understanding and generation.
\newblock In H.~Wallach, H.~Larochelle, A.~Beygelzimer, F.~Alch\'{e}-Buc,
  E.~Fox, and R.~Garnett, editors, \emph{Advances in Neural Information
  Processing Systems 32}, pages 13042--13054. Curran Associates, Inc.

\bibitem[{Dong et~al.(2019{\natexlab{b}})Dong, Li, Rezagholizadeh, and
  Cheung}]{dong-etal-2019-editnts}
Yue Dong, Zichao Li, Mehdi Rezagholizadeh, and Jackie Chi~Kit Cheung.
  2019{\natexlab{b}}.
\newblock \href {https://doi.org/10.18653/v1/P19-1331} {{E}dit{NTS}: An neural
  programmer-interpreter model for sentence simplification through explicit
  editing}.
\newblock In \emph{Proceedings of the 57th Annual Meeting of the Association
  for Computational Linguistics}, pages 3393--3402, Florence, Italy.
  Association for Computational Linguistics.

\bibitem[{Durmus et~al.(2020)Durmus, He, and Diab}]{durmus2020feqa}
Esin Durmus, He~He, and Mona Diab. 2020.
\newblock \href {https://doi.org/10.18653/v1/2020.acl-main.454} {{FEQA}: A
  question answering evaluation framework for faithfulness assessment in
  abstractive summarization}.
\newblock In \emph{Proceedings of the 58th Annual Meeting of the Association
  for Computational Linguistics}, pages 5055--5070, Online. Association for
  Computational Linguistics.

\bibitem[{Dziri et~al.(2019)Dziri, Kamalloo, Mathewson, and
  Zaiane}]{dziri-etal-2019-augmenting}
Nouha Dziri, Ehsan Kamalloo, Kory Mathewson, and Osmar Zaiane. 2019.
\newblock \href {https://doi.org/10.18653/v1/W19-4103} {Augmenting neural
  response generation with context-aware topical attention}.
\newblock In \emph{Proceedings of the First Workshop on NLP for Conversational
  AI}, pages 18--31, Florence, Italy. Association for Computational
  Linguistics.

\bibitem[{Falke et~al.(2019)Falke, Ribeiro, Utama, Dagan, and
  Gurevych}]{falke-etal-2019-ranking}
Tobias Falke, Leonardo F.~R. Ribeiro, Prasetya~Ajie Utama, Ido Dagan, and Iryna
  Gurevych. 2019.
\newblock Ranking generated summaries by correctness: An interesting but
  challenging application for natural language inference.
\newblock In \emph{Proceedings of the 57th Annual Meeting of the Association
  for Computational Linguistics}, pages 2214--2220, Florence, Italy.

\bibitem[{Fan et~al.(2018)Fan, Lewis, and Dauphin}]{fan-etal-2018-hierarchical}
Angela Fan, Mike Lewis, and Yann Dauphin. 2018.
\newblock \href {https://doi.org/10.18653/v1/P18-1082} {Hierarchical neural
  story generation}.
\newblock In \emph{Proceedings of the 56th Annual Meeting of the Association
  for Computational Linguistics (Volume 1: Long Papers)}, pages 889--898,
  Melbourne, Australia. Association for Computational Linguistics.

\bibitem[{Fellbaum(1998)}]{fellbaum98wordnet}
Christiane Fellbaum, editor. 1998.
\newblock \emph{{WordNet: an electronic lexical database}}.
\newblock MIT Press.

\bibitem[{Freitag et~al.(2020)Freitag, Grangier, and
  Caswell}]{freitag-etal-2020-bleu}
Markus Freitag, David Grangier, and Isaac Caswell. 2020.
\newblock \href {https://www.aclweb.org/anthology/2020.emnlp-main.5} {{BLEU}
  might be guilty but references are not innocent}.
\newblock In \emph{Proceedings of the 2020 Conference on Empirical Methods in
  Natural Language Processing (EMNLP)}, pages 61--71, Online. Association for
  Computational Linguistics.

\bibitem[{Gabriel et~al.(2020)Gabriel, Celikyilmaz, Jha, Choi, and
  Gao}]{gabriel2020figure}
Saadia Gabriel, Asli Celikyilmaz, Rahul Jha, Yejin Choi, and Jianfeng Gao.
  2020.
\newblock \href {http://arxiv.org/abs/2010.12834} {Go figure! a meta evaluation
  of factuality in summarization}.

\bibitem[{Gehrmann et~al.(2021)Gehrmann, Adewumi, Aggarwal, Ammanamanchi,
  Anuoluwapo, Bosselut, Chandu, Clinciu, Das, Dhole, Du, Durmus, Dusek, Emezue,
  Gangal, Garbacea, Hashimoto, Hou, Jernite, Jhamtani, Ji, Jolly, Kumar,
  Ladhak, Madaan, Maddela, Mahajan, Mahamood, Majumder, Martins,
  McMillan{-}Major, Mille, van Miltenburg, Nadeem, Narayan, Nikolaev,
  Niyongabo, Osei, Parikh, Perez{-}Beltrachini, Rao, Raunak, Rodriguez,
  Santhanam, Sedoc, Sellam, Shaikh, Shimorina, Cabezudo, Strobelt, Subramani,
  Xu, Yang, Yerukola, and Zhou}]{gem2021}
Sebastian Gehrmann, Tosin~P. Adewumi, Karmanya Aggarwal, Pawan~Sasanka
  Ammanamanchi, Aremu Anuoluwapo, Antoine Bosselut, Khyathi~Raghavi Chandu,
  Miruna{-}Adriana Clinciu, Dipanjan Das, Kaustubh~D. Dhole, Wanyu Du, Esin
  Durmus, Ondrej Dusek, Chris Emezue, Varun Gangal, Cristina Garbacea,
  Tatsunori Hashimoto, Yufang Hou, Yacine Jernite, Harsh Jhamtani, Yangfeng Ji,
  Shailza Jolly, Dhruv Kumar, Faisal Ladhak, Aman Madaan, Mounica Maddela,
  Khyati Mahajan, Saad Mahamood, Bodhisattwa~Prasad Majumder, Pedro~Henrique
  Martins, Angelina McMillan{-}Major, Simon Mille, Emiel van Miltenburg, Moin
  Nadeem, Shashi Narayan, Vitaly Nikolaev, Rubungo~Andre Niyongabo, Salomey
  Osei, Ankur~P. Parikh, Laura Perez{-}Beltrachini, Niranjan~Ramesh Rao, Vikas
  Raunak, Juan~Diego Rodriguez, Sashank Santhanam, Jo{\~{a}}o Sedoc, Thibault
  Sellam, Samira Shaikh, Anastasia Shimorina, Marco Antonio~Sobrevilla
  Cabezudo, Hendrik Strobelt, Nishant Subramani, Wei Xu, Diyi Yang, Akhila
  Yerukola, and Jiawei Zhou. 2021.
\newblock \href {https://arxiv.org/abs/2102.01672} {The {GEM} benchmark:
  Natural language generation, its evaluation and metrics}.
\newblock \emph{CoRR}, abs/2102.01672.

\bibitem[{Gehrmann et~al.(2018)Gehrmann, Deng, and Rush}]{gehrmann-emnlp18}
Sebastian Gehrmann, Yuntian Deng, and Alexander Rush. 2018.
\newblock Bottom-up abstractive summarization.
\newblock In \emph{Proceedings of the 2018 Conference on Empirical Methods in
  Natural Language Processing}, pages 4098--4109, Brussels, Belgium.
  Association for Computational Linguistics.

\bibitem[{Geva et~al.(2019)Geva, Malmi, Szpektor, and Berant}]{discofuse}
Mor Geva, Eric Malmi, Idan Szpektor, and Jonathan Berant. 2019.
\newblock {D}isco{F}use: {A} large-scale dataset for discourse-based sentence
  fusion.
\newblock In \emph{Proceedings of the 2019 Conference of the North {A}merican
  Chapter of the Association for Computational Linguistics: Human Language
  Technologies}, pages 3443--3455, Minneapolis, Minnesota. Association for
  Computational Linguistics.

\bibitem[{Ghosh et~al.(2016)Ghosh, Vinyals, Strope, Roy, Dean, and
  Heck}]{Ghosh2016ContextualL}
Shalini Ghosh, Oriol Vinyals, Brian Strope, Scott Roy, Tom Dean, and Larry
  Heck. 2016.
\newblock Contextual {LSTM (CLSTM)} models for large scale {NLP} tasks.
\newblock \emph{CoRR}, abs/1602.06291.

\bibitem[{Gu et~al.(2016)Gu, Lu, Li, and Li}]{gu-etal-2016-incorporating}
Jiatao Gu, Zhengdong Lu, Hang Li, and Victor~O.K. Li. 2016.
\newblock \href {https://doi.org/10.18653/v1/P16-1154} {Incorporating copying
  mechanism in sequence-to-sequence learning}.
\newblock In \emph{Proceedings of the 54th Annual Meeting of the Association
  for Computational Linguistics (Volume 1: Long Papers)}, pages 1631--1640,
  Berlin, Germany. Association for Computational Linguistics.

\bibitem[{Hashimoto et~al.(2019)Hashimoto, Zhang, and
  Liang}]{hashimoto-etal-2019-unifying}
Tatsunori Hashimoto, Hugh Zhang, and Percy Liang. 2019.
\newblock \href {https://doi.org/10.18653/v1/N19-1169} {Unifying human and
  statistical evaluation for natural language generation}.
\newblock In \emph{Proceedings of the 2019 Conference of the North {A}merican
  Chapter of the Association for Computational Linguistics: Human Language
  Technologies, Volume 1 (Long and Short Papers)}, pages 1689--1701,
  Minneapolis, Minnesota. Association for Computational Linguistics.

\bibitem[{Hermann et~al.(2015)Hermann, Kocisky, Grefenstette, Espeholt, Kay,
  Suleyman, and Blunsom}]{hermann-nips15}
Karl~Moritz Hermann, Tomas Kocisky, Edward Grefenstette, Lasse Espeholt, Will
  Kay, Mustafa Suleyman, and Phil Blunsom. 2015.
\newblock Teaching machines to read and comprehend.
\newblock In C.~Cortes, N.~D. Lawrence, D.~D. Lee, M.~Sugiyama, and R.~Garnett,
  editors, \emph{Advances in Neural Information Processing Systems 28}, pages
  1693--1701. Curran Associates, Inc.

\bibitem[{Hochreiter and Schmidhuber(1997)}]{lstm}
Sepp Hochreiter and J\"{u}rgen Schmidhuber. 1997.
\newblock Long short-term memory.
\newblock \emph{Neural computation}, 9(8):1735--1780.

\bibitem[{Holtzman et~al.(2020)Holtzman, Buys, Du, Forbes, and Choi}]{nucleus}
Ari Holtzman, Jan Buys, Li~Du, Maxwell Forbes, and Yejin Choi. 2020.
\newblock \href {https://openreview.net/forum?id=rygGQyrFvH} {The curious case
  of neural text degeneration}.
\newblock In \emph{International Conference on Learning Representations}.

\bibitem[{Karmaker~Santu et~al.(2019)Karmaker~Santu, Veeramachaneni, and
  Zhai}]{karmaker-santu-etal-2019-tilm}
Shubhra~Kanti Karmaker~Santu, Kalyan Veeramachaneni, and Chengxiang Zhai. 2019.
\newblock \href {https://doi.org/10.18653/v1/K19-1073} {{TILM}: Neural language
  models with evolving topical influence}.
\newblock In \emph{Proceedings of the 23rd Conference on Computational Natural
  Language Learning (CoNLL)}, pages 778--788, Hong Kong, China. Association for
  Computational Linguistics.

\bibitem[{Kim et~al.(2019)Kim, Kim, and Kim}]{Kim:2018:arXiv}
Byeongchang Kim, Hyunwoo Kim, and Gunhee Kim. 2019.
\newblock Abstractive summarization of {R}eddit posts with multi-level memory
  networks.
\newblock In \emph{Proceedings of the 2019 Conference of the North {A}merican
  Chapter of the Association for Computational Linguistics: Human Language
  Technologies}, pages 2519--2531, Minneapolis, Minnesota. Association for
  Computational Linguistics.

\bibitem[{Kryscinski et~al.(2019)Kryscinski, McCann, Xiong, and
  Socher}]{Kryscinski2019EvaluatingTF}
Wojciech Kryscinski, Bryan McCann, Caiming Xiong, and Richard Socher. 2019.
\newblock Evaluating the factual consistency of abstractive text summarization.
\newblock \emph{CoRR}, abs/1910.12840.

\bibitem[{Kryscinski et~al.(2020)Kryscinski, McCann, Xiong, and
  Socher}]{kryscinski-etal-2020-evaluating}
Wojciech Kryscinski, Bryan McCann, Caiming Xiong, and Richard Socher. 2020.
\newblock \href {https://www.aclweb.org/anthology/2020.emnlp-main.750}
  {Evaluating the factual consistency of abstractive text summarization}.
\newblock In \emph{Proceedings of the 2020 Conference on Empirical Methods in
  Natural Language Processing (EMNLP)}, pages 9332--9346, Online. Association
  for Computational Linguistics.

\bibitem[{Kryściński et~al.(2019)Kryściński, McCann, Xiong, and
  Socher}]{factcc}
Wojciech Kryściński, Bryan McCann, Caiming Xiong, and Richard Socher. 2019.
\newblock \href {http://arxiv.org/abs/1910.12840} {Evaluating the factual
  consistency of abstractive text summarization}.

\bibitem[{Kudo and Richardson(2018)}]{sentencepiece}
Taku Kudo and John Richardson. 2018.
\newblock {S}entence{P}iece: {A} simple and language independent subword
  tokenizer and detokenizer for neural text processing.
\newblock In \emph{Proceedings of the 2018 Conference on Empirical Methods in
  Natural Language Processing: System Demonstrations}, pages 66--71, Brussels,
  Belgium. Association for Computational Linguistics.

\bibitem[{Kulikov et~al.(2019)Kulikov, Miller, Cho, and
  Weston}]{kulikov-etal-2019-importance}
Ilia Kulikov, Alexander Miller, Kyunghyun Cho, and Jason Weston. 2019.
\newblock \href {https://doi.org/10.18653/v1/W19-8609} {Importance of search
  and evaluation strategies in neural dialogue modeling}.
\newblock In \emph{Proceedings of the 12th International Conference on Natural
  Language Generation}, pages 76--87, Tokyo, Japan. Association for
  Computational Linguistics.

\bibitem[{Lewis et~al.(2019)Lewis, Liu, Goyal, Ghazvininejad, Mohamed, Levy,
  Stoyanov, and Zettlemoyer}]{bart}
Mike Lewis, Yinhan Liu, Naman Goyal, Marjan Ghazvininejad, Abdelrahman Mohamed,
  Omer Levy, Ves Stoyanov, and Luke Zettlemoyer. 2019.
\newblock {BART: D}enoising sequence-to-sequence pre-training for natural
  language generation, translation, and comprehension.
\newblock \emph{CoRR}, abs/1910.13461.

\bibitem[{Li et~al.(2016{\natexlab{a}})Li, Galley, Brockett, Gao, and
  Dolan}]{li-etal-2016-diversity}
Jiwei Li, Michel Galley, Chris Brockett, Jianfeng Gao, and Bill Dolan.
  2016{\natexlab{a}}.
\newblock \href {https://doi.org/10.18653/v1/N16-1014} {A diversity-promoting
  objective function for neural conversation models}.
\newblock In \emph{Proceedings of the 2016 Conference of the North {A}merican
  Chapter of the Association for Computational Linguistics: Human Language
  Technologies}, pages 110--119, San Diego, California. Association for
  Computational Linguistics.

\bibitem[{Li et~al.(2016{\natexlab{b}})Li, Monroe, and Jurafsky}]{LiMJ16}
Jiwei Li, Will Monroe, and Dan Jurafsky. 2016{\natexlab{b}}.
\newblock \href {http://arxiv.org/abs/1611.08562} {A simple, fast diverse
  decoding algorithm for neural generation}.
\newblock \emph{CoRR}, abs/1611.08562.

\bibitem[{Lin and Hovy(2000)}]{lin-hovy-2000-automated}
Chin-Yew Lin and Eduard Hovy. 2000.
\newblock \href {https://www.aclweb.org/anthology/C00-1072} {The automated
  acquisition of topic signatures for text summarization}.
\newblock In \emph{{COLING} 2000 Volume 1: The 18th International Conference on
  Computational Linguistics}.

\bibitem[{Lin and Hovy(2003)}]{rouge}
Chin~Yew Lin and Eduard Hovy. 2003.
\newblock Automatic evaluation of summaries using n-gram co-occurrence
  statistics.
\newblock In \emph{Proceedings of the 2003 Human Language Technology Conference
  of the North American Chapter of the Association for Computational
  Linguistics}, pages 150--157.

\bibitem[{Liu et~al.(2019)Liu, Ott, Goyal, Du, Joshi, Chen, Levy, Lewis,
  Zettlemoyer, and Stoyanov}]{roberta}
Yinhan Liu, Myle Ott, Naman Goyal, Jingfei Du, Mandar Joshi, Danqi Chen, Omer
  Levy, Mike Lewis, Luke Zettlemoyer, and Veselin Stoyanov. 2019.
\newblock {RoBERTa: A} robustly optimized {BERT} pretraining approach.
\newblock \emph{CoRR}, abs/1907.11692.

\bibitem[{Mallinson et~al.(2020)Mallinson, Severyn, Malmi, and
  Garrido}]{mallinson2020felix}
Jonathan Mallinson, Aliaksei Severyn, Eric Malmi, and Guillermo Garrido. 2020.
\newblock \href {http://arxiv.org/abs/2003.10687} {Felix: Flexible text editing
  through tagging and insertion}.

\bibitem[{Malmi et~al.(2019)Malmi, Krause, Rothe, Mirylenka, and
  Severyn}]{malmi-etal-2019-encode}
Eric Malmi, Sebastian Krause, Sascha Rothe, Daniil Mirylenka, and Aliaksei
  Severyn. 2019.
\newblock \href {https://doi.org/10.18653/v1/D19-1510} {Encode, tag, realize:
  High-precision text editing}.
\newblock In \emph{Proceedings of the 2019 Conference on Empirical Methods in
  Natural Language Processing and the 9th International Joint Conference on
  Natural Language Processing (EMNLP-IJCNLP)}, pages 5054--5065, Hong Kong,
  China. Association for Computational Linguistics.

\bibitem[{Mani(2001)}]{mani2001automatic}
Inderjeet Mani. 2001.
\newblock \emph{Automatic summarization}, volume~3.
\newblock John Benjamins Publishing.

\bibitem[{Martins et~al.(2020)Martins, Marinho, and
  Martins}]{martins-etal-2020-sparse}
Pedro~Henrique Martins, Zita Marinho, and Andr{\'e} F.~T. Martins. 2020.
\newblock \href {https://www.aclweb.org/anthology/2020.emnlp-main.348} {Sparse
  text generation}.
\newblock In \emph{Proceedings of the 2020 Conference on Empirical Methods in
  Natural Language Processing (EMNLP)}, pages 4252--4273, Online. Association
  for Computational Linguistics.

\bibitem[{Maynez et~al.(2020)Maynez, Narayan, Bohnet, and
  McDonald}]{maynez2020faithfulness}
Joshua Maynez, Shashi Narayan, Bernd Bohnet, and Ryan McDonald. 2020.
\newblock \href {https://doi.org/10.18653/v1/2020.acl-main.173} {On
  faithfulness and factuality in abstractive summarization}.
\newblock In \emph{Proceedings of the 58th Annual Meeting of the Association
  for Computational Linguistics}, pages 1906--1919, Online. Association for
  Computational Linguistics.

\bibitem[{Mikolov and Zweig(2012)}]{mikolovZ12}
Tomas Mikolov and Geoffrey Zweig. 2012.
\newblock Context dependent recurrent neural network language model.
\newblock In \emph{Proceedings of the Spoken Language Technology Workshop},
  pages 234--239. IEEE.

\bibitem[{Narayan et~al.(2018)Narayan, Cohen, and
  Lapata}]{narayan-etal-2018-xsum}
Shashi Narayan, Shay~B. Cohen, and Mirella Lapata. 2018.
\newblock Don{'}t give me the details, just the summary! topic-aware
  convolutional neural networks for extreme summarization.
\newblock In \emph{Proceedings of the 2018 Conference on Empirical Methods in
  Natural Language Processing}, pages 1797--1807, Brussels, Belgium.
  Association for Computational Linguistics.

\bibitem[{Narayan et~al.(2016)Narayan, Reddy, and
  Cohen}]{narayan-etal-2016-paraphrase}
Shashi Narayan, Siva Reddy, and Shay~B. Cohen. 2016.
\newblock \href {https://doi.org/10.18653/v1/W16-6625} {Paraphrase generation
  from latent-variable {PCFG}s for semantic parsing}.
\newblock In \emph{Proceedings of the 9th International Natural Language
  Generation conference}, pages 153--162, Edinburgh, UK. Association for
  Computational Linguistics.

\bibitem[{Nenkova and McKeown(2011)}]{Nenkova:McKeown:2011}
Ani Nenkova and Kathleen McKeown. 2011.
\newblock Automatic summarization.
\newblock \emph{Foundations and Trends in Information Retrieval},
  5(2--3):103--233.

\bibitem[{Pasunuru and Bansal(2018)}]{Pasunuru-multireward18}
Ramakanth Pasunuru and Mohit Bansal. 2018.
\newblock Multi-reward reinforced summarization with saliency and entailment.
\newblock In \emph{Proceedings of the 2018 Conference of the North American
  Chapter of the Association for Computational Linguistics: Human Language
  Technologies}, pages 646--653. Association for Computational Linguistics.

\bibitem[{Qi et~al.(2020)Qi, Yan, Gong, Liu, Duan, Chen, Zhang, and
  Zhou}]{qi-etal-2020-prophetnet}
Weizhen Qi, Yu~Yan, Yeyun Gong, Dayiheng Liu, Nan Duan, Jiusheng Chen, Ruofei
  Zhang, and Ming Zhou. 2020.
\newblock \href {https://www.aclweb.org/anthology/2020.findings-emnlp.217}
  {{P}rophet{N}et: Predicting future n-gram for
  sequence-to-{S}equence{P}re-training}.
\newblock In \emph{Findings of the Association for Computational Linguistics:
  EMNLP 2020}, pages 2401--2410, Online. Association for Computational
  Linguistics.

\bibitem[{Radford et~al.(2018)Radford, Narasimhan, Salimans, and
  Sutskever}]{gpt}
Alec Radford, Karthik Narasimhan, Tim Salimans, and Ilya Sutskever. 2018.
\newblock Improving language understanding by generative pre-training.
\newblock \emph{Technical report, OpenAI}.

\bibitem[{Raffel et~al.(2019)Raffel, Shazeer, Roberts, Lee, Narang, Matena,
  Zhou, Li, and Liu}]{t5}
Colin Raffel, Noam Shazeer, Adam Roberts, Katherine Lee, Sharan Narang, Michael
  Matena, Yanqi Zhou, Wei Li, and Peter~J. Liu. 2019.
\newblock Exploring the limits of transfer learning with a unified text-to-text
  transformer.
\newblock \emph{CoRR}, abs/1901.07291.

\bibitem[{Rajpurkar et~al.(2018)Rajpurkar, Jia, and Liang}]{squad}
Pranav Rajpurkar, Robin Jia, and Percy Liang. 2018.
\newblock Know what you don{'}t know: Unanswerable questions for {SQ}u{AD}.
\newblock In \emph{Proceedings of the 56th Annual Meeting of the Association
  for Computational Linguistics}, pages 784--789, Melbourne, Australia.
  Association for Computational Linguistics.

\bibitem[{Rei et~al.(2020)Rei, Stewart, Farinha, and Lavie}]{comet}
Ricardo Rei, Craig Stewart, Ana~C Farinha, and Alon Lavie. 2020.
\newblock \href {https://doi.org/10.18653/v1/2020.emnlp-main.213} {{COMET}: A
  neural framework for {MT} evaluation}.
\newblock In \emph{Proceedings of the 2020 Conference on Empirical Methods in
  Natural Language Processing (EMNLP)}, pages 2685--2702, Online. Association
  for Computational Linguistics.

\bibitem[{Rothe et~al.(2020)Rothe, Narayan, and Severyn}]{rothe2020leveraging}
Sascha Rothe, Shashi Narayan, and Aliaksei Severyn. 2020.
\newblock Leveraging pre-trained checkpoints for sequence generation tasks.
\newblock \emph{Transactions of the Association for Computational Linguistics},
  8:264--280.

\bibitem[{Schoch et~al.(2020)Schoch, Yang, and Ji}]{schoch-etal-2020-problem}
Stephanie Schoch, Diyi Yang, and Yangfeng Ji. 2020.
\newblock \href {https://www.aclweb.org/anthology/2020.evalnlgeval-1.2}
  {{``}this is a problem, don{'}t you agree?{''} framing and bias in human
  evaluation for natural language generation}.
\newblock In \emph{Proceedings of the 1st Workshop on Evaluating NLG
  Evaluation}, pages 10--16, Online (Dublin, Ireland). Association for
  Computational Linguistics.

\bibitem[{See et~al.(2017)See, Liu, and Manning}]{see-acl17}
Abigail See, Peter~J. Liu, and Christopher~D. Manning. 2017.
\newblock Get to the point: Summarization with pointer-generator networks.
\newblock In \emph{Proceedings of the 55th Annual Meeting of the Association
  for Computational Linguistics}, pages 1073--1083, Vancouver, Canada.
  Association for Computational Linguistics.

\bibitem[{Sellam et~al.(2020)Sellam, Das, and Parikh}]{bleurt}
Thibault Sellam, Dipanjan Das, and Ankur Parikh. 2020.
\newblock \href {https://doi.org/10.18653/v1/2020.acl-main.704} {{BLEURT}:
  Learning robust metrics for text generation}.
\newblock In \emph{Proceedings of the 58th Annual Meeting of the Association
  for Computational Linguistics}, pages 7881--7892, Online. Association for
  Computational Linguistics.

\bibitem[{Song et~al.(2020)Song, Lebanoff, Guo, Qiu, Xue, Li, Yu, and
  Liu}]{Song2020JointPA}
Kaiqiang Song, Logan Lebanoff, Q.~Guo, Xipeng Qiu, X.~Xue, Chen Li, Dong Yu,
  and Fei Liu. 2020.
\newblock Joint parsing and generation for abstractive summarization.
\newblock In \emph{AAAI}.

\bibitem[{Song et~al.(2019)Song, Tan, Qin, Lu, and Liu}]{mass_icml19}
Kaitao Song, Xu~Tan, Tao Qin, Jianfeng Lu, and Tie{-}Yan Liu. 2019.
\newblock {MASS: M}asked sequence to sequence pre-training for language
  generation.
\newblock In \emph{Proceedings of the 36th International Conference on Machine
  Learning}, volume~97, pages 5926--5936. PMLR.

\bibitem[{Sultan et~al.(2020)Sultan, Chandel, Fernandez~Astudillo, and
  Castelli}]{sultan-etal-2020-importance}
Md~Arafat Sultan, Shubham Chandel, Ram{\'o}n Fernandez~Astudillo, and Vittorio
  Castelli. 2020.
\newblock \href {https://doi.org/10.18653/v1/2020.acl-main.500} {On the
  importance of diversity in question generation for {QA}}.
\newblock In \emph{Proceedings of the 58th Annual Meeting of the Association
  for Computational Linguistics}, pages 5651--5656, Online. Association for
  Computational Linguistics.

\bibitem[{Tian et~al.(2019)Tian, Narayan, Sellam, and
  Parikh}]{tian2019sticking}
Ran Tian, Shashi Narayan, Thibault Sellam, and Ankur~P. Parikh. 2019.
\newblock \href {http://arxiv.org/abs/1910.08684} {Sticking to the facts:
  Confident decoding for faithful data-to-text generation}.

\bibitem[{Vaswani et~al.(2017)Vaswani, Shazeer, Parmar, Uszkoreit, Jones,
  Gomez, Kaiser, and Polosukhin}]{transformer}
Ashish Vaswani, Noam Shazeer, Niki Parmar, Jakob Uszkoreit, Llion Jones,
  Aidan~N Gomez, Lukasz Kaiser, and Illia Polosukhin. 2017.
\newblock Attention is all you need.
\newblock In \emph{Advances in Neural Information Processing Systems 30}, pages
  5998--6008.

\bibitem[{Vijayakumar et~al.(2018)Vijayakumar, Cogswell, Selvaraju, Sun, Lee,
  Crandall, and Batra}]{diversebeam}
Ashwin~K. Vijayakumar, Michael Cogswell, Ramprasaath~R. Selvaraju, Qing Sun,
  Stefan Lee, David~J. Crandall, and Dhruv Batra. 2018.
\newblock \href
  {https://www.aaai.org/ocs/index.php/AAAI/AAAI18/paper/view/17329} {Diverse
  beam search for improved description of complex scenes}.
\newblock In \emph{Proceedings of the Thirty-Second {AAAI} Conference on
  Artificial Intelligence, (AAAI-18), the 30th innovative Applications of
  Artificial Intelligence (IAAI-18), and the 8th {AAAI} Symposium on
  Educational Advances in Artificial Intelligence (EAAI-18), New Orleans,
  Louisiana, USA, February 2-7, 2018}, pages 7371--7379. {AAAI} Press.

\bibitem[{Wang et~al.(2020{\natexlab{a}})Wang, Cho, and
  Lewis}]{Wang2020AskingAA}
Alex Wang, Kyunghyun Cho, and Mike Lewis. 2020{\natexlab{a}}.
\newblock \href {https://doi.org/10.18653/v1/2020.acl-main.450} {Asking and
  answering questions to evaluate the factual consistency of summaries}.
\newblock In \emph{Proceedings of the 58th Annual Meeting of the Association
  for Computational Linguistics}, pages 5008--5020, Online. Association for
  Computational Linguistics.

\bibitem[{Wang et~al.(2020{\natexlab{b}})Wang, Rao, Zhang, Qin, Tian, and
  Wang}]{wang-etal-2020-diversify}
Zhen Wang, Siwei Rao, Jie Zhang, Zhen Qin, Guangjian Tian, and Jun Wang.
  2020{\natexlab{b}}.
\newblock \href {https://www.aclweb.org/anthology/2020.findings-emnlp.194}
  {Diversify question generation with continuous content selectors and question
  type modeling}.
\newblock In \emph{Findings of the Association for Computational Linguistics:
  EMNLP 2020}, pages 2134--2143, Online. Association for Computational
  Linguistics.

\bibitem[{Wang et~al.(2020{\natexlab{c}})Wang, Duan, Zhang, Wang, Tian, Chen,
  and Zhou}]{wang-etal-2020-friendly}
Zhengjue Wang, Zhibin Duan, Hao Zhang, Chaojie Wang, Long Tian, Bo~Chen, and
  Mingyuan Zhou. 2020{\natexlab{c}}.
\newblock \href {https://www.aclweb.org/anthology/2020.emnlp-main.35} {Friendly
  topic assistant for transformer based abstractive summarization}.
\newblock In \emph{Proceedings of the 2020 Conference on Empirical Methods in
  Natural Language Processing (EMNLP)}, pages 485--497, Online. Association for
  Computational Linguistics.

\bibitem[{Wang et~al.(2020{\natexlab{d}})Wang, Wang, An, Yu, and
  Chen}]{wang2020faithful}
Zhenyi Wang, Xiaoyang Wang, Bang An, Dong Yu, and Changyou Chen.
  2020{\natexlab{d}}.
\newblock \href {http://arxiv.org/abs/2005.00969} {Towards faithful neural
  table-to-text generation with content-matching constraints}.

\bibitem[{Welleck et~al.(2019{\natexlab{a}})Welleck, Kulikov, Roller, Dinan,
  Cho, and Weston}]{Welleck2019}
Sean Welleck, Ilia Kulikov, Stephen Roller, Emily Dinan, Kyunghyun Cho, and
  Jason Weston. 2019{\natexlab{a}}.
\newblock \href {http://arxiv.org/abs/1908.04319} {Neural text generation with
  unlikelihood training}.
\newblock \emph{CoRR}, abs/1908.04319.

\bibitem[{Welleck et~al.(2019{\natexlab{b}})Welleck, Weston, Szlam, and
  Cho}]{welleck-etal-2019-dialogue}
Sean Welleck, Jason Weston, Arthur Szlam, and Kyunghyun Cho.
  2019{\natexlab{b}}.
\newblock Dialogue natural language inference.
\newblock In \emph{Proceedings of the 57th Annual Meeting of the Association
  for Computational Linguistics}, pages 3731--3741, Florence, Italy.

\bibitem[{Williams et~al.(2018)Williams, Nangia, and Bowman}]{mnli}
Adina Williams, Nikita Nangia, and Samuel Bowman. 2018.
\newblock A broad-coverage challenge corpus for sentence understanding through
  inference.
\newblock In \emph{Proceedings of the 2018 Conference of the North American
  Chapter of the Association for Computational Linguistics: Human Language
  Technologies}, pages 1112--1122. Association for Computational Linguistics.

\bibitem[{Xing et~al.(2017)Xing, Wu, Wu, Liu, Huang, Zhou, and
  Ma}]{topicgen-aaai17}
Chen Xing, Wei Wu, Yu~Wu, Jie Liu, Yalou Huang, Ming Zhou, and Wei-Ying Ma.
  2017.
\newblock Topic aware neural response generation.
\newblock In \emph{Proceedings of the Thirty-First AAAI Conference on
  Artificial Intelligence}, AAAI'17, page 3351–3357. AAAI Press.

\bibitem[{Xu et~al.(2018)Xu, Ren, Lin, and Sun}]{xu-etal-2018-diversity}
Jingjing Xu, Xuancheng Ren, Junyang Lin, and Xu~Sun. 2018.
\newblock \href {https://doi.org/10.18653/v1/D18-1428} {Diversity-promoting
  {GAN}: A cross-entropy based generative adversarial network for diversified
  text generation}.
\newblock In \emph{Proceedings of the 2018 Conference on Empirical Methods in
  Natural Language Processing}, pages 3940--3949, Brussels, Belgium.
  Association for Computational Linguistics.

\bibitem[{Xu et~al.(2020)Xu, Li, Yuan, Wu, He, and Zhou}]{xu-etal-2020-self}
Song Xu, Haoran Li, Peng Yuan, Youzheng Wu, Xiaodong He, and Bowen Zhou. 2020.
\newblock \href {https://doi.org/10.18653/v1/2020.acl-main.125} {Self-attention
  guided copy mechanism for abstractive summarization}.
\newblock In \emph{Proceedings of the 58th Annual Meeting of the Association
  for Computational Linguistics}, pages 1355--1362, Online. Association for
  Computational Linguistics.

\bibitem[{Xu et~al.(2016)Xu, Napoles, Pavlick, Chen, and
  Callison-Burch}]{xu-etal-2016-optimizing}
Wei Xu, Courtney Napoles, Ellie Pavlick, Quanze Chen, and Chris Callison-Burch.
  2016.
\newblock Optimizing statistical machine translation for text simplification.
\newblock \emph{Transactions of the Association for Computational Linguistics},
  4:401--415.

\bibitem[{Yang et~al.(2019)Yang, Dai, Yang, Carbonell, Salakhutdinov, and
  Le}]{xlnet_arxiv19}
Zhilin Yang, Zihang Dai, Yiming Yang, Jaime~G. Carbonell, Ruslan Salakhutdinov,
  and Quoc~V. Le. 2019.
\newblock {XLNet: G}eneralized autoregressive pretraining for language
  understanding.
\newblock \emph{CoRR}, abs/1906.08237.

\bibitem[{Zhang et~al.(2019)Zhang, Zhao, Saleh, and Liu}]{zhang2019pegasus}
Jingqing Zhang, Yao Zhao, Mohammad Saleh, and Peter~J. Liu. 2019.
\newblock \href {http://arxiv.org/abs/1912.08777} {Pegasus: Pre-training with
  extracted gap-sentences for abstractive summarization}.

\bibitem[{Zhang et~al.(2020)Zhang, Kishore, Wu, Weinberger, and
  Artzi}]{bertscore}
Tianyi Zhang, Varsha Kishore, Felix Wu, Kilian~Q. Weinberger, and Yoav Artzi.
  2020.
\newblock {BERTScore: E}valuating text generation with {BERT}.
\newblock In \emph{Proceedings of the 8th International Conference on Learning
  Representations}, Virtual Conference, Formerly Addis Ababa Ethiopia.

\bibitem[{Zhou et~al.(2012)Zhou, Hannah, Dunson, and Carin}]{pmlr-v22-zhou12c}
Mingyuan Zhou, Lauren Hannah, David Dunson, and Lawrence Carin. 2012.
\newblock \href {http://proceedings.mlr.press/v22/zhou12c.html} {Beta-negative
  binomial process and poisson factor analysis}.
\newblock In \emph{Proceedings of the Fifteenth International Conference on
  Artificial Intelligence and Statistics}, volume~22 of \emph{Proceedings of
  Machine Learning Research}, pages 1462--1471, La Palma, Canary Islands. PMLR.

\bibitem[{Zhou et~al.(2017)Zhou, Yang, Wei, and
  Zhou}]{zhou-etal-2017-selective}
Qingyu Zhou, Nan Yang, Furu Wei, and Ming Zhou. 2017.
\newblock \href {https://doi.org/10.18653/v1/P17-1101} {Selective encoding for
  abstractive sentence summarization}.
\newblock In \emph{Proceedings of the 55th Annual Meeting of the Association
  for Computational Linguistics (Volume 1: Long Papers)}, pages 1095--1104,
  Vancouver, Canada. Association for Computational Linguistics.

\end{thebibliography}
\bibliographystyle{acl_natbib}

\clearpage

\appendix

\section{Implementation and Reproducibility Details}
\label{sec:impl}

Following \newcite{rothe2020leveraging}, the encoder and decoder of \roberta and \famer models are initialized with public RoBERTa checkpoints. The encoder and decoder parameters are shared in both cases. Only the encoder-decoder attention parameters are initialized randomly. For \famer, the focus attention parameters are also randomly initialized. We experiment with large RoBERTa checkpoints with 24 layers, a hidden size of 1024, filter size of 4096, 16 attention heads, and a vocabulary with 50K sentence pieces \cite{sentencepiece}. \roberta has around 455M parameters and \famer has 463M parameters, with an additional 8M parameters. Our \pegasus and \famep implementation also have the same configuration, except for the encoder-decoder attention parameters which are pretrained. 

We used Cloud TPU v3 accelerators for training. All models are fine-tuned on the target task using Adam with a learning rate of 0.05. We use a linear learning rate warm up with 40k steps, normalized by the square root of the hidden size, and a square root decay. We do not perform any tuning on these hyperparameters. We use a global batch size of 128 document-summary pairs. We adapt to different number of training steps depending on the training data sizes. Models are trained for 400k and 200k steps for \cnndm and \xsum respectively, saving check-points every 1000 steps. We choose the best model based on \rougel performance on the respective validation set.

The vocabulary for functional tokens $F$ is constructed by taking the most frequent sentence pieces in the training set. We tune $|F|$ using the respective validation sets; for \xsum, we choose $f=500$ frequent sentence pieces and for \cnndm, $f=1000$. For all our experiments with the \fame models, the beam size is set to 4. 

We use Cloud TPU v3 accelerators for computing entailment scores which takes about 20 minutes for the two datasets' test sets. Question generation and answering for Feqa are run on a NVIDIA V100 GPU, and it takes between 8-12 hours for one setting of each test set.

\begin{table}[th!]
\centering
\scriptsize %
\begin{tabular}{r|ccc}
\toprule
\multirow{2}{*}{Models} & \multicolumn{3}{c}{\cnndm} \\
& R1 & R2 & RL \\
\midrule
Lead & 39.60 & 17.70 & 36.20 \\
PtGen \cite{see-acl17} & 39.53 & 17.28 &  36.38  \\ 
Bottom-Up \cite{gehrmann-emnlp18} & 41.22 & 18.68 & 38.34 \\ 
SAGCopy \cite{xu-etal-2020-self} & 42.53 & 19.92 & 39.44 \\
MASS \cite{mass_icml19} &  42.12 & 19.50 & 39.01 \\
UniLM \cite{unilm_arxiv19} & 43.33 & 20.21 & 40.51 \\
BART \cite{bart} & 44.16 & 21.28 & 40.90 \\
T5 \cite{t5} & 43.52 & \textbf{\underline{21.55}} & 40.69 \\
\pegasus (C4, \citeauthor{zhang2019pegasus}, \citeyear{zhang2019pegasus}) & 43.90 & 21.20 & 40.76 \\
\pegasus (HugeNews, \citeauthor{zhang2019pegasus}, \citeyear{zhang2019pegasus}) & 44.17 & 21.47 & 41.11 \\
ProphetNet \cite{qi-etal-2020-prophetnet} & \textbf{\underline{44.20}} &  21.17 & \textbf{\underline{41.30}} \\
\midrule
\roberta \cite{rothe2020leveraging} & 39.88 & 18.66 & 37.22 \\ 
\famer (ours) & 40.27 & 18.43 & 37.51 \\
\pegasus  (ours) & 42.62 & 20.38 & 39.61 \\
\famep  (ours) & \textbf{42.95} & \textbf{20.79} & \textbf{39.90} \\
\bottomrule
\end{tabular}
\caption{Abstractive summarization results on \cnndm datasets. The \textbf{\underline{underlined bold}} results are from the best performing models from literature and the \textbf{bold} results are the best performing \fame models.}
\label{tab:cnn-results-complete}
\vspace{-0.1cm}
\end{table}

\begin{table*}[t!]
\centering
\footnotesize
\begin{tabular}{ r | ccc|cc|cc|c}
\toprule
\multirow{2}{*}{Models} & \multirow{2}{*}{Len.} & Rep. & R1(P\%) & \multicolumn{2}{c|}{doc. $\rightarrow$ sum.} &  \multicolumn{2}{c|}{Feqa} & \multirow{2}{*}{BERTSc.} \\
&  & \% & With doc. & ent. ($\uparrow$) & $\neg$ cont. & acc. & avg.(\#Q) & \\
\midrule
\roberta & 52.1 & 77.6 & 92.7 & 88.8 & 96.4 & 37.3 & 18.1 & 76.0 \\
\famer & 55.5 & 79.6 & 92.5 & 87.3 & 96.3 & 35.2 & 19.3 & 76.1 \\
\pegasus & 58.1 & 69.4 & 95.0 & 90.9 & 97.5 & 40.3 & 21.0 & 76.8 \\
\famep & 58.5 & 71.0 & 95.3 & \textbf{91.0} & \textbf{97.6} & \textbf{41.1} & 21.1 & \textbf{76.9} \\
\bottomrule
\end{tabular}
\vspace{-0.1cm}
\caption{Faithfulness and qualitative assessment of summaries on \cnndm dataset.}
\label{tab:cnn-quality-scores}
\vspace{-0.1cm}
\end{table*}

\section{Abstractive Summarization Results on CNN/DailyMail}
\label{sec:cnn-results}

The \cnndm dataset \cite{hermann-nips15} consists of 287,227/13,368/11,490 training/validation/test document-summary pairs. The \cnndm summaries are in the form of bullet-point story highlights and exhibit a high degree of extraction, requiring the models to learn to copy from the source documents. The \xsum summaries, on the other hand, are extreme, in that the documents are summarized into single-sentence summaries with a high level of abstractiveness. For comparison, the \xsum summaries show a much larger percentages of novel constructions than found in \cnndm summaries ($35.8 / 83.5 / 95.5 / 98.5$ vs $16.8 / 54.3 / 72.4 / 80.4$ novel $1 /  2  / 3 / 4$-grams). We use the original cased version. During training, the input documents are truncated to 512 tokens and the length of the summaries are limited to 128 tokens. 

Table~\ref{tab:cnn-results-complete} and~\ref{tab:cnn-quality-scores} present complete results for \cnndm dataset. We see similar kind of improvements as observed in Table~\ref{tab:xsum-results-full}, except for \rougetwo for \famer which is 0.23 points worse than the \roberta baseline. 
Our best model \famep performs better than both copy mechanism models: LSTM-based PtGen \cite{see-acl17} and Transformer-based SAGCopy \cite{xu-etal-2020-self}. \famep performs worse when compared with T5 \cite{t5}, the original \pegasus \cite{zhang2019pegasus} and ProphetNet \cite{qi-etal-2020-prophetnet}. This can be expected as the number of parameters in \famep is almost half of T5 or ProphetNet, and is 100M less than that in the original \pegasus.

\famer performs worse than \roberta on both ent.\ and Feqa measures for \cnndm, similar to \rougetwo in Table~\ref{tab:cnn-results-complete}.  We hypothesize that this is due to the extractive nature of the \cnndm dataset and the fact that it is not able to copy tokens from the input to the necessary extent as the encoder-decoder attention is not pre-trained. Moreover, Feqa scores for \roberta and \famer may not be fully comparable due to variation in their summary lengths and the number of Feqa questions generated; the \famer summaries, on average, are 3 words longer and generate 1.2 more questions than that of \roberta.
Nevertheless, we don't see this kind of drop in $\neg$cont. scores (i.e., summary not contradicting, either entailed by or neutral to the document) and BERTScores.

\section{Text Editing Results}
\label{sec:texted-results}
We also train the \fame models on two text editing tasks: (i) for sentence fusion -- the problem of combining multiple sentences into a single coherent sentence -- we used the ``balanced Wikipedia'' portion of the DiscoFuse dataset \citep{discofuse}, and (ii) for split-and-rephrase -- the reverse  task  of  sentence  fusion -- we used the WikiSplit dataset \citep{wikisplit}, which consists of 1M examples of sentence splits extracted from the Wikipedia edit history. As the name suggests, both text editing tasks require a low degree of abstraction.

For both the tasks, we train the models for 300k steps with a global batch size of 256. The input and output are padded to a length of 128, which covers 100\% of the training, evaluation and test data. The vocabulary for functional tokens $F$ is constructed by taking the top 100 and 500 sentence pieces for DiscoFuse and WikiSplit respectively.

\begin{table}[t!]
\centering
\scriptsize %
\begin{tabular}{r|ccc}
\toprule
DiscoFuse & Exact  & SARI & BLEU \\
\midrule
\cite{discofuse}   & 51.1          & 84.5 & --\\
LaserTagger \cite{malmi-etal-2019-encode} & 53.8 & 85.5 & -- \\
Felix \cite{mallinson2020felix} & 61.3 & 88.8 & -- \\ 
\roberta \cite{rothe2020leveraging} & 66.6 & 90.3 & --  \\
\pegasus (ours) & \underline{67.4} & \underline{90.5} & 95.8 \\ 
\famep (ours) & \textbf{67.8} & \textbf{90.7} & \textbf{95.9} \\
\midrule
 WikiSplit & Exact  & SARI   & BLEU \\
\midrule
\cite{wikisplit} & 14.3        & 61.5       & 76.4  \\
LaseTagger \cite{malmi-etal-2019-encode} & 15.2 & 61.7 & 76.3 \\ 
\roberta \cite{rothe2020leveraging} & 16.4 & 63.8   &  \textbf{77.4}  \\
\pegasus (ours) & \underline{16.6} & \textbf{64.1} & \textbf{77.4} \\
\famep (ours) & \textbf{16.8} & \textbf{64.1} & 77.3 \\
\bottomrule
\end{tabular}
\caption{Text editing results on Discofuse and WikiSplit. The \underline{underlined} scores beat the current state-of-the-art and the \textbf{bold} scores are the new state-of-the-art.}
\label{tab:textedit}
\vspace{-0.4cm}
\end{table}

We report corpus-level BLEU\footnote{We use NLTK v3.2.2 with case sensitive scoring to estimate BLEU scores.}, the exact match accuracy, and SARI scores \citep{xu-etal-2016-optimizing}\footnote{\label{sari_implementation}SARI 
is a lexical similarity metric which compares the model's output to multiple references and the input in order to assess the model's ability to add, delete, and keep an $n$-gram. It's implementation is available at: \url{https://github.com/tensorflow/tensor2tensor/blob/master/tensor2tensor/utils/sari_hook.py}.}. The results can be seen in Table~\ref{tab:textedit}. The vanilla \pegasus model already beats the current state-of-the-art on both DiscoFuse and WikiSplit. The \famep model performs better, albeit by a small margin, on all metrics on DiscoFuse. On WikiSplit, it has a higher exact match accuracy while maintaining the SARI score and performs 0.1 BLEU worse than \pegasus.

\section{Controlled Generation with focus attention using Top-$k$ tokens}
\label{sec:controlled}

Table~\ref{tab:focus-sampling-controlled} presents results from our controlled summary generation experiments with top-$k$ tokens from $t_X$ using focus attention (\ftopk) on the \xsum test set. In Figures \ref{fig:topk-sentpiece} and \ref{fig:topk-rouge1}, we describe how \famer consistently outperforms \famep at lower values of $k \in \{50,100,200,500,1000\}$ due to their peaky and smooth $t_X$, respectively. While Figure~\ref{fig:topk-rouge1} only plots \rouge-1 F1 scores, %
Table~\ref{tab:focus-sampling-controlled} additionally reports \rouge-2, \rouge-L, entailment, Feqa, and BERTScores. Figure~\ref{fig:fame-controlled-predictions} presents predictions from models using \ftopk for the article presented in Figures~\ref{fig:intro-fame-predictions} and \ref{fig:fame-article-predictions}.

\section{Diverse Summarization with \divtopk, \divnucleus and \fsamplek}
\label{sec:diverse}

Figures~\ref{fig:fame-predictions-fsample}, \ref{fig:roberta-predictions-topk}, \ref{fig:roberta-predictions-nucleus}, \ref{fig:pegasus-predictions-topk} and \ref{fig:pegasus-predictions-nucleus} show the diverse summaries generated using \fsamplek, \divtopk and \divnucleus sampling methods for the article shown in Figure~\ref{fig:fame-article-predictions}.

\begin{table*}[t!]
\centering
\footnotesize %
\begin{tabular}{r|cccccc}
\toprule
\multirow{2}{*}{Metrics} & \multicolumn{3}{c}{\rouge} & \multirow{2}{*}{ent.} & \multirow{2}{*}{Feqa} & \multirow{2}{*}{BERTScore} \\
& R1 & R2 & RL & \\
\midrule
\roberta & 41.45 & 18.79 & 33.90 & 39.1 & 19.8 & 80.6 \\
\famer & \textbf{42.15} & \textbf{19.68} & \textbf{34.81} & \textbf{41.3} & \textbf{21.2} & \textbf{80.8} \\
\famer ($\mathrm{Focus}_{\mathrm{top},k=50}$) & 30.90 & 10.60 & 24.85 & 27.1 & 10.6 & 74.2 \\
\famer ($\mathrm{Focus}_{\mathrm{top},k=100}$) & 33.62 & 12.39 & 27.14 & 30.3 & 12.4 & 74.2 \\
\famer ($\mathrm{Focus}_{\mathrm{top},k=200}$) & 35.99 & 14.12 & 29.23 & 32.4 & 13.9 & 77.3  \\
\famer ($\mathrm{Focus}_{\mathrm{top},k=500}$) & 38.29 & 16.04 & 31.30 & 35.8 & 15.9 & 78.6  \\
\famer ($\mathrm{Focus}_{\mathrm{top},k=1000}$) & 39.58 & 17.18 & 32.49 & 37.3 & 17.3 & 79.3 \\
\famer ($\mathrm{Focus}_{\mathrm{top},k=10000}$) & 41.58 & 19.13 & 34.30 & 40.7 & 20.2 & 80.5 \\
\midrule
\pegasus & 44.85 & 22.26 & 37.03 & 43.6 & 24.5 & 81.7 \\
\famep & \textbf{45.31} & \textbf{22.75} & \textbf{37.46} & \textbf{44.8} & \textbf{24.8} & \textbf{81.9} \\
\famep ($\mathrm{Focus}_{\mathrm{top},k=50}$) & 24.30 & 7.52 & 19.32 & 20.8 & 8.0 & 68.8 \\
\famep ($\mathrm{Focus}_{\mathrm{top},k=100}$) & 27.77 & 9.26 & 22.09 & 24.1 & 9.3 & 71.3 \\
\famep ($\mathrm{Focus}_{\mathrm{top},k=200}$) & 31.05 & 11.14 & 24.82 & 27.0 & 10.8 & 73.6 \\
\famep ($\mathrm{Focus}_{\mathrm{top},k=500}$) & 34.99 & 13.65 & 28.19 & 31.0 & 13.0 & 76.2  \\
\famep ($\mathrm{Focus}_{\mathrm{top},k=1000}$) & 37.40 & 15.30 & 30.16 & 33.6 & 14.9 & 75.9 \\
\famep ($\mathrm{Focus}_{\mathrm{top},k=10000}$) & 42.76 & 19.89 & 34.97 & 40.2 & 20.1 & 80.5 \\
\bottomrule
\end{tabular}
\caption{Assessment of controlled summary generation with focus sampling \ftopk on the \xsum test set. 
We experiment with limiting \fame models to different sizes of vocabulary $V_k$ using the topic distribution $t_X$; in particular, we experiment with $k=\{50, 100, 200, 500, 1000, 10000\}$. We also report numbers for \roberta, \famer, \pegasus and \famep, using the whole vocabulary of size 50k. 
The \textbf{bold} results in each block are the best performing \roberta-based and \pegasus-based models.}
\label{tab:focus-sampling-controlled}
\end{table*}

\begin{figure*}[t!]
  \center{\footnotesize %
    \begin{tabular}{l p{12cm}}
    \toprule 
    \textbf{\gold} & Australia has expelled an Israeli diplomat saying Israel was behind the forging of Australian passports linked to the murder of a Hamas operative in Dubai. \\
    \midrule
    \textbf{Article} & Australia's foreign minister said these were ``not the actions of a friend''. \\
    & The UK took similar action in March, after concluding that Israel was responsible for the use of forged UK passports in the plot. \\
    & The Israeli foreign ministry said Australia's decision was disappointing. \\
    & Ministry spokesman Yigal Palmor said it was ``not in line with the importance and the quality of the relationship between our countries''. \\
    & 'Sorrow not anger' \\
    & At least four forged Australian passports were used in the killing of Mahmoud al-Mabhouh in Dubai in January. The originals belonged to Australians living in Israel. \\
    & The Australian government said a police investigation had left it in no doubt that the Israeli authorities were behind ``the abuse and counterfeiting of the passports''. \\
    & As a result Foreign Minister Stephen Smith asked Israel to withdraw a diplomat, whom he did not identify. \\
    & ``The decision to ask Israel to remove from Australia one of its officers at the Israeli embassy in Canberra is not something which fills the Australian government with any joy,'' he said. \\
    & ``On the contrary, the decision was made much more in sorrow than in anger.'' \\
    & Passports from France, Ireland, Germany and Britain were used in the operation, and in March, the British government expelled an Israeli diplomat from London. \\
    & The Israeli government has said there is no proof that it was behind the killing, although Dubai officials have said they are 99.9\% sure that agents from Mossad were responsible. \\
    \midrule
    \textbf{\roberta} & Australia has asked \textcolor{orangered}{Australia} to withdraw an Israeli diplomat from its embassy in Canberra after an alleged plot to kill a \textcolor{orangered}{Abu Dhabi militant} in Dubai. \\
    \textbf{\famer} & Australia has asked Israel to withdraw one of its diplomats from its embassy in Canberra after \textcolor{orangered}{it admitted} it used forged passports. \\
    \textbf{\pegasus} & Australia has expelled an Israeli diplomat after concluding that forged Australian passports used in the killing of a Hamas militant in Dubai were issued by Israel. \\
    \textbf{\famep} & The Australian government has expelled an Israeli diplomat over the use of forged Australian passports in the killing of a Hamas militant in Dubai. \\ 
    \bottomrule
    \end{tabular}     
  }
  \caption{A 2010 BBC article from the \xsum testset, its human written summary and model predictions from \roberta, and \pegasus, with and without \fame. The text in \textcolor{orangered}{orange} is not supported by the input article.}
  \label{fig:fame-article-predictions}
\end{figure*}

\begin{figure*}[t!]
  \center{\footnotesize %
    \begin{tabular}{l p{11cm}}
    \toprule 
    \textbf{\famer ($\mathrm{Focus}_{\mathrm{top},k=50}$)} & Australia has said \textcolor{orangered}{it will not be expelled an ambassador from Australia following the alleged s agent for the so-called Arab Arab State.} \\
    \textbf{\famer ($\mathrm{Focus}_{\mathrm{top},k=100}$)} & Australia has said \textcolor{orangered}{it will not be expelled an ambassador from Australia} following the killing of a terror agent in the Arab world. \\
    \textbf{\famer ($\mathrm{Focus}_{\mathrm{top},k=200}$)} & Australia has said \textcolor{orangered}{it will not be expelled an ambassador from Australia following the killing of an Australian terror suspect} in the Arab world. \\
    \textbf{\famer ($\mathrm{Focus}_{\mathrm{top},k=500}$)} &  Australia has asked Israel \textcolor{orangered}{to end its diplomatic investigation into an alleged plot to murder an Australian terror suspect}. \\
    \textbf{\famer ($\mathrm{Focus}_{\mathrm{top},k=1000}$)} & Australia has asked Israel to strip an ambassador from its embassy following the death of an Arab man in Dubai. \\
    \textbf{\famer ($\mathrm{Focus}_{\mathrm{top},k=10000}$)} & Australia has asked Israel to withdraw one of its diplomats from its embassy in Canberra following the death of a terror suspect. \\ 
    \midrule
    \textbf{\famep ($\mathrm{Focus}_{\mathrm{top},k=50}$)} & \textcolor{orangered}{The Israeli government has been expelled} from the country after it was found that the country's security agency, the Israeli intelligence agency, was to be \textcolor{orangered}{to be found to have used a number of the country's out-of-country p when it was used in the Emirates car-j best.} \\
    \textbf{\famep ($\mathrm{Focus}_{\mathrm{top},k=100}$)} & \textcolor{orangered}{The Israeli government has been expelled} from the country after it was found that the country's security agency, the Israeli intelligence agency, had used the country's visas in the \textcolor{orangered}{Emirates terror}. \\
    \textbf{\famep ($\mathrm{Focus}_{\mathrm{top},k=200}$)} & The Australian government has expelled an Israeli diplomats after it found that the country's security agency, the Israeli intelligence agency, had used the country's visas in the \textcolor{orangered}{Emirates terror attack}. \\
    \textbf{\famep ($\mathrm{Focus}_{\mathrm{top},k=500}$)} & The Australian government has expelled an Israeli diplomatic staff after accusing the country's security agency, the Israeli intelligence agency, of using a number of Australian visas in the \textcolor{orangered}{Emirates terror attack}. \\
    \textbf{\famep ($\mathrm{Focus}_{\mathrm{top},k=1000}$)} & Australia has expelled an Israeli diplomatic staff after accusing the country's security agency, the Israeli military's intelligence agency, of being responsible for the use of Australian visas used in the killing of a Palestinian. \\
    \textbf{\famep ($\mathrm{Focus}_{\mathrm{top},k=10000}$)} & Australia has expelled an Israeli diplomat over the use of forged Australian passports in the killing of a Hamas militant in Dubai. \\ 
    \bottomrule
    \end{tabular}     
  }
  \caption{Model predictions with focus sampling \ftopk, a controlled generation setting. The text in \textcolor{orangered}{orange} is not supported by the input article. We note that with smaller values of $k$, both \roberta-based and \pegasus-based models tend to hallucinate more often.}
  \label{fig:fame-controlled-predictions}
\end{figure*}

\begin{figure*}[t!]
  \center{\footnotesize %
    \begin{tabular}{p{15cm}}
    \toprule 
    \textbf{\famer (\fsamplek)} \\
    Australia has asked Israel to strip one of its diplomats from its embassy following the death of an Arab man in Dubai. \\
    Australia has \textcolor{orangered}{asked Israel to end its diplomatic investigation} into an alleged plot to murder an Australian terror suspect. \\
    Australia has asked Israel to strip one of its diplomats from its embassy in Australia over the death of a terror suspect. \\
    \midrule
    \textbf{\famep (\fsamplek)} \\
    The Australian government has expelled an Israeli diplomatic staff after accusing it of using a number of Australian visas in the killing of a Palestinian \textcolor{orangered}{in a car bombing}. \\
    The Australian government has expelled an Israeli diplomatic staff after it said the country was responsible for the use of Australian visas used in the killing of a Palestinian \textcolor{orangered}{in a car bombing}. \\
    Australia has expelled an Israeli diplomatic staff after accusing the country's security agency, the Israeli military's intelligence agency, of being responsible for the use of Australian visas used in the killing of a Palestinian. \\
    Australia has expelled an Israeli diplomatic mission after accusing the country's security agency, the Israeli military's intelligence agency, of being responsible for the use of Australian visas used in the killing of a Palestinian in the Arab city of Emirates.\\
    The Australian government has expelled an Israeli diplomatic staff after it said the country was responsible for the use of Australian visas used in the killing of a Palestinian in the Middle East.\\
    \bottomrule
    \end{tabular}     
  }
  \caption{\fame model predictions with \fsamplek ($k=10000$). The text in \textcolor{orangered}{orange} is not supported by the input article.\label{fig:fame-predictions-fsample}}
\end{figure*}

\begin{figure*}[t!]
  \center{\footnotesize %
    \begin{tabular}{p{15cm}}
    \toprule 
    \textbf{\roberta (\divtopk)} \\
    Australia has asked for an Ivan ``shivers'' officer to be asked to leave Australia after the performance of an Israeli flag was alleged to have been used as terrorism suspects in Dubai.\\ 
    Australia has asked an Israeli ambassador to Sydney over an alleged implicated Australian diplomat alleging the murder of a Australian national in Dubai.\\
    Israel has asked Israel to withdraw an Israeli ambassador from Canberra amid claims that the alleged invasion of its territory by a foreign agent was behind the murder of a terror suspect in Abuabad.\\
    Australia has asked Israel to withdraw a diplomat Izzy Kanhuh, an Israeli diplomat involved in solving tensions over the sale of imported shotguns for the Dubai Abu Dhabiuddin bombing.\\
    Australia has asked Australia to withdraw an ambassador from the country, amid a growing row over the alleged role of an Israel-based Abu Abu Malak director of agents.\\
    Australia has asked Israel to replace its ambassador over a fatal stabbing in Sydney last week.\\
    Australia has asked Israel to withdraw an Egyptian diplomat following the suicide of a suspected Abu Abu Mabhulas in the Australian capital, Canberra.\\
    Australia has asked Australia for an official withdrawal from its embassy in Sydney after the death of a Palestinian diplomat in a Dublin diplomatic fanbase earlier this month.\\
    Australia has asked Israel to withdraw an Israeli diplomat as part of a probe into the alleged involvement in the murder of a Abu Abuab militant.\\
    Australia has asked an Israeli diplomat to be withdrawn from the country over the Diamondad bombing of a Abu Waduh as part of an investigation into its 2002 murders of a Abu Abu Baye bomber.\\
    \midrule
    \textbf{\famer (\divtopk)} \\
    Australia has played down claims its state ambassador was involved in finding out why the Mossad spy agent was behind the Rio stabbing.\\
    Australia has asked Israel to withdraw one of its diplomats after it confessed the so-called Mossad agent agent had used a fake Melbourne funery.\\
    Australia says it will withdraw an envoy after the Israel spy agent accused of involvement in the murder of an Arab smuggler was suspended.\\
    Australia has asked Israel to expel one of its citizens after the country leaked the state agent that led later a deadly mafia murder in Dubai.\\
    Australia has asked Israel to withdraw its consulate at Canberra because from its embassy after it claimed it used the Falcon fuelling plan for a suicide bomb.\\
    Australia has asked Israel to withdraw its support for Europe's embassy for its arrest of an Edinburgh diplomat over the death of a heroin smuggling gang.\\
    Australia has asked Israel to remove an ambassador from its embassy over the shooting dead of an Australian man on a Dubai delivery scheme.\\
    Australia is to withdraw a diplomat from its embassy in Canberra over allegations it worked on the mastermind for an alleged spying plot for the Mossad operation.\\
    Australiachas asked Israel to withdraw an anonymous diplomat from its embassy following investigation into the passage of a Falcon recruiting device.\\
    Australia has asked the Israeli embassy to pull out of its alleged response to the murder of a British terror suspect, accusing it of responsibility.\\
    \midrule
    \textbf{\famer (\fsamplek, \divtopk)} \\
    Australia has asked Israel to answer the decision to honour its state ambassador following the alleged involvement in the killing of a Dubai terror suspect.\\
    Australia has asked Israel for a second diplomat to be expelled from Australia after an alleged plot to murder a man in a bomb plot linked to Mossad.\\
    Australia has asked Israel to make a state diplomat its top diplomat after an alleged plot to bomb an Arab Emirates terror operation was blamed on a terror agent.\\
    Australia has asked Egypt to end its diplomatic at-top diplomatic response to the murder of a top Arab diplomat in the Arab world.\\
    Australia has asked Israel to be expelled from the embassy in Australia following the death of a Sydney spy in a spy investigation.\\
    Australia has asked Israel to to strip an diplomat of its consulate from its embassy since a deadly operation against the Mossad spy agent at a terror squad in Australia last month.\\
    Australia has asked the Israel embassy to withdrawing its diplomats following the death of an Arab man by Mossad agents.\\
    Australia has asked Israel to end the original accusations that a diplomat is responsible for the killing of an agent from Mossad.\\
    Australia has asked Israel to answer the investigation that admitted its diplomats used his agent as a suicide bomb in a Dubai plot.\\
    Australia has asked Israel to support its ambassador after it admitted being involved in the murder of a suspect in the deadly one-off terror killing in a Melbourne bomb attack.\\
    \bottomrule
    \end{tabular}     
  }
  \caption{Diverse summaries predicted using \roberta and \famer models with \divtopk.\label{fig:roberta-predictions-topk}}
\end{figure*}

\begin{figure*}[t!]
  \center{\footnotesize %
    \begin{tabular}{p{15cm}}
    \toprule 
    \textbf{\roberta (\divnucleus)} \\
    Australia says man hasenzelled an Israeli envoy following the arrest of one of its diplomats in Dubai from the countries' deepest-running terrorism resistant group. badly documented.\\
    Australia has asked for Israel to out retrieving an Israeli diplomat who was expelled from the country after Australia accused the FBI of involvement in a 2013 murder in rogue Myersad drug smuggling operation.\\
    Australia has asked Israel to remove an envoy from its embassy in Sydney in an escalating row over the killing of a Yazad Bin Ab alcohol dealer in the United Arab Emirates.\\
    Australia has asked Israel to clarify its response to a data breach cull from rendition with a relapse of a suspected Abu Abuabuded jihadist.\\
    Australia has asked Australia to pull out of Israel after an Israeli diplomat was accused of having used sreleased Australian agent Abuadab in the murder of an Abu Dhabi carrier.\\
    Australia Herb Allen has led Australia's ambassadorsaints over an investigation into what was allegedly led by one of its diplomats at Nessadab consultancy in Dubai.\\
    Australia has urged Israel to withdraw an ambassador pshorze over alleged links to the murder of a Sydney binnington.\\
    Australia has asked the Israeli ambassador to Australia over an inferno at a Sydney diplomatic consulate for a senior recruiter which printers had wanted a Willis bin Laden agent to be charged.\\
    Australia has asked Australia for the withdrawal of an Israeli ambassador after an investigation into it was linked to a Vietnam-based gang in which a young dungeonsad spy was killed.\\
    Australia has asked Israel for an emotional withdrawal from its embassy in Canberra, accusing an Israeli diplomat of involvement in a feuding plot to kill a terror suspect.\\
    \midrule
    \textbf{\famer (\divnucleus)} \\
    Australia has asked Israel to withdraw an Israeli official over a Team Mossad bomb plot that left one of its suspects in the Dubai Arab desert. \\
    Australia has asked all Israeli diplomats to leave Canberra after the living place of an alleged Russian special forces agent was identified at the email bug held bymacadad.\\
    Australia is to withdraw an official sensitivity inquiry from its foreign ministers after Israel was accused of involvement in a plot to kill a Dubai terror suspect.\\
    Australia has asked Israel deep back into allegations it carried out a wanted plot Cunning deaths in a Dubai plot by Mossad agents.\\
    AustraliaplayedAX has asked the Israeli government to withdraw an official language envoy from its embassies following the killing of a murdered cons consulate officer.\\
    Australia has asked an Israeli official to withdraw an official ambassador after it made a murder in a deadly shooting Presumably by Mossad.\\
    Australia has asked Israel over allegations that an agent used forged passports to plot the Woolstroken murder by agentsbased in Pakistan.\\
    Australia has asked Israeli authorities to withdraw an official diplomat from Australia after the mafia was accused by the Israel embassy of contributing to its alleged failed murder of an Alquer Arab Shia terrorist.\\
    Australia has asked an Israeli embassy to withdraw a diplomat from Australia following the Jewlands' murder of an unnamed man.\\
    Australia has annexed its embassy up tolishes at the start of the year after Israel confirmed it assessed the role of an undercover officer during the Dubai heroinmer plot.\\
    \midrule
    \textbf{\famer (\fsamplek, \divnucleus)} \\
    Australia has asked Israel to expelled an embassy diplomat over a deadly Sydney plot to spy on the Mossad operation. \\
    Australia has asked Israel to end its diplomatic inruru from Australia after it accused its diplomatic staff of involvement in last year's deadly attack on a Melbourne terror attack.\\
    Israel has asked Israel to make an embassy ambassador over a deadly email killing of a man in a terror plot.\\
    Australia has asked Israel to strip its diplomatic staff of its passport following an alleged plot to murder a Dubai terror suspect.\\
    Israel has asked Israel to expelled one of its diplomats after the Mossad agent accused a Melbourne man of being the agent for the Mossad spy agent for his role in an alleged plot to murder a man.\\
    Australia has asked Israel to strip a top envoy from his embassy following its investigation into the killing of an alleged spy in a Melbourne email plot.\\
    Australia has asked Israel to expelled one diplomat following allegations it used a military agent to spy for Mossad.\\
    Australia has asked the Israel embassy to be expelled from Australia after an Australian diplomat was found guilty of his role in the murder of an Australian terror suspect.\\
    Australia is to expelled its top diplomat from Australia after his country was accused by the UN of being responsible for an alleged plot to murder a Melbourne-Arab m intelligence agent.\\
    Australia has asked Israel to strip an ambassador from its embassy, in response to the death of a Sydney-from-agent for the so-called ``Mossad, was responsible''.\\
    \bottomrule
    \end{tabular}     
  }
  \caption{Diverse summaries predicted using \roberta and \famer models with \divnucleus.\label{fig:roberta-predictions-nucleus}}
\end{figure*}

\begin{figure*}[t!]
  \center{\footnotesize %
    \begin{tabular}{p{15cm}}
    \toprule 
    \textbf{\pegasus (\divtopk)} \\
    Australia has expelled an Israeli diplomat over the use of forged Australian passports in the killing of Hamas detainee Mahmoud al-Mabhouh in Dubai. \\
    Israel has summoned the Australian ambassador to complain after the Australian government said forged passports used in the killing of a Hamas operative in Dubai belonged to Netanyahu's foreign ministry.\\
    The Australian government has ordered Israel to withdraw an officer over the use of forged Australian passports used by the 2013 murder of a Lebanese opposition figure in Dubai.\\
    The Australian government has expelled an Israeli diplomat over allegations that fake Australian passports were used to kill a Lebanese militant in Egypt two years ago.\\
    Australia has asked Israel to withdraw a diplomat over the use of forged Australian passports to kill a Hamas operative in January.\\
    Australia has expelled an Israeli diplomat in a row over the authenticated use of forged Australian passports in last year's killing of a Hamas figure in Dubai.\\
    Australia says it is expulsion an Israeli diplomat in protest over Israel's alleged role in the killing of a Hamas militant in Dubai.\\
    Australia has recalled a diplomat from Israel after accusing Berlin of fabricating false passports used in the assassination of a Hamas operative in Dubai.\\
    Israel has been asked to withdraw an official from Australia, accusing it of complicity in the falsification of Australian passports used in the killing of a Hamas operative in Dubai in January.\\
    Israel has withdrawn one of its diplomats after Canberra said it concluded that Passport Bureau agents participated in an internal Mossad plot to kill a Hamas operative in Dubai.\\
    \midrule
    \textbf{\famep (\divtopk)} \\
    Australia has expelled an Israeli diplomat after it concluded somebody close to Israel's security agency, Mossad, owned forged passports which were used to abduct a Hamas rocket maker.\\
    Australia has expelled an Israeli diplomat over allegations that its intelligence agency Mossad was behind the use of forged passports in the killing of a suspected Palestinian militant.\\
    Australia has expelled an Israeli diplomat amid accusations Israel-run Mossad used forged Australian passports in the killing of a Hamas militant.\\
    Australia has expelled an Israeli diplomat in a dispute over the use of stolen Australian passports for a hit in the Dubai killing of a Lebanese militant earlier this year.\\
    An Israeli diplomat has been expelled from Australia after a Sydney police team concluded that agents from the country's security agency Mossad took part in the poisoning of Egypt's president.\\
    The Australian government has expelled an Israeli diplomat, after it concluded that his desk was responsible for the issuance of forged Australian passports used in the killing of a Hamas militant.\\
    Australia has recalled her envoy from Israel, after finding that an Israeli diplomat was responsible for the counterfeiting of passports used by the Unesco agency director who was killed in Dubai.\\
    The Australian government has asked Israel to withdraw from its Embassy in Melbourne after accusing it of using forged Australian passports to fund the killing of a Palestinian militant.\\
    The Australian government has asked Israel to withdraw its ambassador for failing to acknowledge its role in the use of forged Australian passports in the killing of a British businessman.\\
    Australia has formally demanded the removal of anIsrael diplomat in response to a decision to accuse the Jewish organisation Mossad of use of forged Australian passports mentioned in a Dubai bombing plot.\\
    \midrule
    \textbf{\famep (\fsamplek, \divtopk)} \\
    Australia has expelled an Israeli because ``its anti-espionage agents'' used visas from other nations to issue a Palestinian agent's body £2.3m (£``2.3,1) car and land lift to Hezbollah in the killing of a senior Palestinian in the city: \\
    The Australian government has ejected an Israeli at its embassy over the use of Australia's visas in the killing of a terror attack in the city of D'scale.\\
    Australia has expelled an Israeli diplomats following an investigation into the use of the country's travel services as cards used in a terror attack.\\
    Australia has expelled an Israeli embassy transport staff after a police investigation found the country's intelligence agency, the Israeli intelligence agency, was at responsible.\\
    Australia has expelled an Israeli embassy gathering after it said the country was responsible for the use of the use of Australia's Australian emails in an emailed attack on a former Australian consulate in the Arab world.\\
    Australia has expelled an Israeli diplomatic side after accusing it of using at first issued Australian DNA test cards to produce the Irish agent in the stepped-up Emirates bombing.\\
    Australia has expelled Israeli diplomats after its foreign minister said the country had been to be responsible for the use of staged Australian emails by Israeli intelligence.\\
    Australia has told Israel to withdraw a diplomatic mission from its country - after it said it was ``in no Petroleum to Finish'' the killing of a Palestinian in the killing of a terror the network by the Israeli security agency, enzymes.\\
    Australia has demanded Israel withdraw a  diplomats following the Israel Security Service's use of Israeli-issued Australian travel visas to help one of its agents commit a terror attack.\\
    \bottomrule
    \end{tabular}     
  }
  \caption{Diverse summaries predicted using \pegasus and \famep models with \divtopk.\label{fig:pegasus-predictions-topk}}
\end{figure*}

\begin{figure*}[t!]
  \center{\footnotesize %
    \begin{tabular}{p{15cm}}
    \toprule 
    \textbf{\pegasus (\divnucleus)} \\
    Israel hasracuse withdrawn an envoy after the Australian government said it concluded that Israeli agents used forged passports used to kill a Dubai Bendigo businessman.\\
    Australia has demanded the withdrawal of an Israeli diplomat, saying his arrival in Canberra was only necessary to deal with a spillover from the killing of a Hamas militant in Dubai in January.\\
    The Australian government has recalled an Israeli diplomat over accusation that fake Australian passports used 436 kilometres (300 miles) from Canberra in the death of a Hamas militant were stolen by Israeli agents.\\
    The Australian government has recalled aLatis from Israel for having strong evidence that their embassy was used to counterfeite passports used in the killing of a bidder in Dubai.\\
    Australia has expelled an Israeli diplomat in a row about the use429 Australian passportsocally used in the killing of an intends in Dubai. slew of passports were used cosmetics in the killing.\\
    Australia is seeking to expel an Israeli envoyrolet over the use of forged Australian passports in the murder of a militant in Dubaiselection.\\
    Australia has expelled an Israeli diplomat after saying it was ``certain'' eagle-eyed undercover agents were Rhys Shapiro and Glenn Clift, who used forged Australian passports to kill an Soros geneticist in Dubai in 2015.\\
    Australia has removed the Israeli ambassador following a decision to conclude that forged Australian passports used in the death of a Palestinian Deals in the DesertDex were collaborated from Israel.\\
    Australia has recalled an Israeli diplomat, accusing Tel Aviv of ``engaging in a pattern of alarming behaviour'', after concluding that forged Australian passports were used in the killing of a Hamas operative in Dubai.\\
    Israel has expelled one of its diplomats because of allegations that it helped Isabel al-Mabhouh, a British-based PalestinianORED to be kill in January, by using forged Australian passports.\\
    \midrule
    \textbf{\famep (\divnucleus)} \\
    Australia has summoned Idair Kernatic, a Jerusalem consulate official, inv summoned after the extraction of a document touting the use of forged passports for a deadly bomb plot. \\
    Australia has recalled a diplomat from Israel, claiming Israel stole the original identities of passports used to kill a Hamas operative.\\
    Australia has asked Israel Fever to withdraw a diplomat after New Zealand said Israeli agents used the fake local passports used to identify a keyoine Killer.\\
    The Australian government has asked Israel to withdraw a diplomat after claiming the Jewish terror group Mossad used forged Australian passports in a plot to murder a Dubai imam.\\
    Australia has expelled an Israeli diplomat after confirming fake Australian passports were used to help the killing ofmagazine boss Mahmoud al-Mabhouh in Dubai.\\
    Australia has withdrawn a military characteristic of Israel after alleging its officials were behind the use of stolen Australian passportsxiety in a Dubai cash-in-transit plot.\\
    Australia has expel an Israeli diplomat over allegations that the country's Mossad spy was behind at least Jong-Bam's Becket murder.\\
    Australia has withdrawn an Israeli diplomat halves its embassy in Canberra over accusations the country's security service, Mossad, was responsible for issuing forged Australian passports.\\
    Australia has asked Israel to withdraw one of its diplomats from Canberra after finding that phony Australian passports were used to kill an Egyptian cleric.\\
    Australia has asked Israel to withdraw a diplomat after it said Israel was behinduse parts of forged Australian passports used in the bombing of a kayaker in Dubai.\\
    \midrule
    \textbf{\famep (\fsamplek, \divnucleus)} \\
    Australia has expelled an Israeli government agent after accusing it of using the use of Australian travel visas to help Israel's intelligence agency, arrest a Palestinian in a drug operation.\\
    Australia has expelled an Israeli pulled over the use of Australian espionage proteins in a terror attack.\\
    The Australian government says Israel should withdraw a senior police mission from its embassy following an investigation into the use of Australian gel-making equipment in the killing of a Palestinian in a car bombing in a Emirates airport.\\
    Australia has expelled an Israeli diplomats for its support for a Palestinian that was used to hack the email messages of the former head of the intelligence agency, reasoning that the expulsion was ``in the best security'' of the two countries.\\
    Australia has expelled Israel's second in service special operations, after accusing the country's intelligence agency, theahl, of a ``poisoning''.\\
    Australia has expelled an Israeli government in protest at ``the use'' of an Australian denied diplomatic entry in a diplomatic killing in the Arab city of controversives.\\
    Australia has expelled an Israeli posting at its embassy in a ``diplomatic action'', after it was found that Israeli agents had used issued Australian visas in the killing of an Egyptian man.\\
    Australia has expelled an Israeli diplomatic, accusing it of using a home-grown Palestinian with a crime on the plane he was using to be arrested in the West.\\
    The Australian government has expel an Israeli diplomatic team, following a public investigation into the use of Australian visas to help the killing of a Palestinian in the Emirates.\\
    The Australian government has expelled an Israeli embassy consulate in Australia after saying it was ``left in no suggestions'' it was responsible for the use of Australian terror attack credentials in the killing of a Palestinian in the desert.\\
    \bottomrule
    \end{tabular}     
  }
  \caption{Diverse summaries predicted using \pegasus and \famep  models with \divnucleus.\label{fig:pegasus-predictions-nucleus}}
\end{figure*}

\end{document}